# From Transformers to Large Language Models: A systematic review of AI applications in the energy sector towards Agentic Digital Twins

**Gabriel Antonesi[1], Tudor Cioara[1*], Ionut Anghel[1], Vasilis Michalakopoulos[2], Elissaios Sarmas[2], Liana Toderean[1]**

[1]Distributed Systems Research Laboratory, Computer Science Department, Technical University of Cluj-Napoca, G. Barițiu 26-28, 400027 Cluj-Napoca, Romania; gabriel.antonesi@cs.utcluj.ro, ionut.anghel@cs.utcluj.ro.

[2]Decision Support Systems Laboratory, School of Electrical & Computer Engineering, National Technical University of Athens, Ir. Politechniou 9, 157 73 Athens, Greece; vmichalakopoulos@epu.ntua.gr, esarmas@epu.ntua.gr.

*Corresponding author: tudor.cioara@cs.utcluj.ro

**Abstract:** Artificial intelligence (AI) has long promised to improve energy management in smart grids by enhancing situational awareness and supporting more effective decision-making. While traditional machine learning has demonstrated notable results in forecasting and optimization, it often struggles with generalization, situational awareness, and heterogeneous data integration. Recent advances in foundation models such as Transformer architecture and Large Language Models (LLMs) have demonstrated improved capabilities in modelling complex temporal and contextual relationships, as well as in multi-modal data fusion which is essential for most AI applications in the energy sector. In this review we synthesize the rapid expanding field of AI applications in the energy domain focusing on Transformers and LLMs. We examine the architectural foundations, domain-specific adaptations and practical implementations of transformer models across various forecasting and grid management tasks. We then explore the emerging role of LLMs in the field: adaptation and fine tuning for the energy sector, the type of tasks they are suited for, and the new challenges they introduce. Along the way, we highlight practical implementations, innovations, and areas where the research frontier is rapidly expanding. These recent developments reviewed underscore a broader trend: Generative AI (GenAI) is beginning to augment decision-making not only in high-level planning but also in day-to-day operations, from forecasting and grid balancing to workforce training and asset onboarding. Building on these developments, we introduce the concept of the *Agentic Digital Twin,* a next-generation model that integrates LLMs to bring autonomy, proactivity, and social interaction into digital twin-based energy management systems. We present the transformational impact of LLMs to each phase of a digital twin and identify the open challenges that need to be addressed for their efficient and effective integration.

## 1. Introduction

In recent years, the energy sector has found itself at the intersection of rapid technological change (advances in grid digitization, smart metering infrastructure, edge computing, and real-time monitoring systems) [1-3] and rising environmental and operational challenges (including the variability introduced by large-scale integration of renewable energy sources, the need to reduce greenhouse gas emissions, and the pressures of decarbonization deadlines) [4-5]. As renewable generation, distributed energy resources, and real-time grid operations become the norm, energy systems are generating vast amounts of complex, multi-scale data [6]. Extracting meaningful, timely insights from this data—whether for forecasting day-ahead solar output or managing demand spikes across a smart grid—is not just a technical challenge. It is a prerequisite for maintaining stability, reliability, and efficiency in an increasingly complex and dynamic energy landscape.

Artificial intelligence (AI) has long promised to enhance decision-making across the energy value chain [7]. And for a time, traditional machine learning methods delivered on that promise in many areas: forecasting electricity demand, predicting equipment failures, optimizing grid operations, and more [8-9]. But as the complexity and dynamism of modern energy systems continue to grow, so too do the limits of these conventional tools. Many struggle to scale, generalize across conditions, or integrate heterogeneous sources like weather data, satellite imagery, and market reports. Recent breakthroughs in foundation models in particular the Transformer architectures and Large Language Models (LLMs) have demonstrated improved generalization capabilities to address above challenges. Moreover, their ability to model complex temporal and contextual relationships as well as the multi-modal reasoning presents an unprecedented opportunity for creating more intelligent applications in the energy sector.

The arrival of transformer-based models marked a turning point. Originally developed for language tasks, transformers brought a fundamentally different way of modelling sequences—relying on self-attention rather than recurrence [10]. In doing so, they offered a more flexible, scalable, and accurate approach to time-series modelling, making them a natural fit for energy forecasting tasks with seasonal, periodic, or highly non-linear patterns. Researchers quickly adapted them to handle load prediction, renewable generation, and even spatio-temporal grid data—with impressive results [11-14].

Nowadays we are witnessing the next phase of this evolution: the rise of large language models (LLMs). These models—like GPT [15], PaLM [16], and domain-specific variants such as TimeGPT [17] or PatchTST [18]—build upon the transformer foundation but scale it up dramatically, training on massive datasets and exhibiting capabilities far beyond language. LLMs can perform zero-shot or few-shot learning, synthesize insights from multiple data types, and adapt to new tasks with minimal retraining. In energy applications, this opens the door to richer, more adaptable and contextualized execution of tasks: fusion of structured and text-based data for improved situational awareness, seamless integration of eternal data sources and conversational capabilities, code generation and energy management actions planning and validation.

But with this potential comes a new set of questions. What are the real advantages of these models in the context of several energy problems like energy forecasting, grid optimization, strategic bidding in different energy markets, and grid/storage/ Renewable Energy Systems (RES) investment planning among others? How do they compare to traditional or even transformer-based methods in terms of performance, interpretability, and operational readiness? Can LLMs be fine-tuned effectively for energy problems, given the domain's unique constraints—data sparsity, regulatory risk, and real-time demands?

This review takes a close look and synthesizes recent advances in AI applications in the energy domain focusing on transformers, and LLMs. We begin with a discussion of transformer-based methods: their architecture, domain-specific adaptations, and real-world applications. We then explore how LLMs are entering the field, the kinds of tasks they are suited for, and the new challenges they introduce. Along the way, we highlight practical implementations, innovations, and areas where the research frontier is rapidly expanding. These recent developments reviewed underscore a broader trend: GenAI is beginning to augment decision-making not only in high-level planning but also in day-to-day operations, from fault detection and grid balancing to workforce training and asset onboarding. As these tools move from proof-of-concept to production, it becomes critical to assess their real-world performance, scalability, and integration challenges within complex energy ecosystems.

Moreover, energy management solutions in smart grid are aiming towards system-level reusable and configurable Digital Twins (DTs) which intelligently combine and merge physical and data-based modelling of individual energy assets components within the energy system. With the advent of LLM models, we believe that the smart grid management will transition towards Agentic Digital Twin that combines data-driven mirroring and real-time simulation of energy system models, with the autonomy, proactivity, and social capabilities of an Intelligent Agent. By integrating LMMs throughout all the phases of a DT supporting individual process they will be transformed into agentic elements able not only to manage energy assets components but also able to interact and cooperate with other DTs into a cohesive structure bridging the gap between theoretical and practical applications of AI in the management of the energy grid. So, instead of a passive twin that just reflects the energy asset or system characteristics, we will develop agentic DT that can act on that data, make decisions, communicate, and even negotiate with other DTs. Therefore, in the discussion section we introduce the concept of Agentic Digital Twin as the next step in energy management presenting the transformational value brought by LLM to each phase of a DT and the open challenges that need to be addressed for their efficient and effective integration.

The paper is structured as follows: Section 2 outlines the foundational elements of transformer architectures for predictive tasks, as well as LLMs basic fine-tuning process for energy management tasks adaptation. Section 3 presents methodological details of the systematic literature review process, including search strategy, inclusion/exclusion criteria, and data sources. Section 4 offers an in-depth synthesis of transformer applications across multiple energy forecasting tasks such as household and building-level demand, renewable energy, integrated energy systems, and specialized decision support tasks like EV charging. Section 5 transitions to LLMs, reviewing their recent applications in energy prediction, efficiency optimization, and smart grid management, including examples of domain-specific adaptations, transfer learning, and generative scenario modelling. Section 6 presents the perspective and challenges introducing the concept of Agentic DT as promising management solutions in smart grid, supported by the adoption of LLMs opening promising directions for future investigation.

## 2. Transformers and LLMs overview

At the core of each transformer model lies the self-attention mechanism, that is used to compute weighted relationships between all pairs of input time steps. By converting inputs into queries, keys, and values, the model calculates attention scores that allow it to dynamically determine which past monitored values are most relevant to forecasting the future, rather than relying solely on fixed-length memory. By design, self-attention proves efficient in capturing long-range temporal dependencies, a very important advantage in prediction applications with seasonal patterns or long memory such as yearly renewable energy patterns or monthly load demand cycles. Instead of using a single attention mechanism, transformers use multi-head attention, where multiple attention heads learn different representations of the inputs and their inter-dependences. Parallel attention leads to a better representation of consumption or generation profiles than single-headed models. Because self-attention alone is permutation-invariant (i.e. insensitive to sequence order), transformers use positional encoding to inject chronological information, to preserve the in-place sequential nature of the data. In practice, common methods such as adding fixed sinusoidal position vectors to the inputs to indicate their time index, or learned positional embeddings are used. This makes sure that a transformer recognizes, for instance, which input corresponds to peak vs. off-peak times, preserving the temporal structure needed for forecasting. Improved encoding methods such computing differences in positions between queries and keys, lead to better learning of local information and long-range dependencies. Spatial-temporal encodings also involve the combination of information from different sources, such as geographic coordinates or sensor placements combined with time-of-day, seasonal cycles, or other time-specific markers. Together, these core concepts – self-attention, multi-head parallelism, and positional encodings – allow traditional, vanilla transformers to capture temporal dynamics in energy time series that earlier architectures struggled with. Additionally, transformer models are, at base, typically organized into an encoder–decoder architecture. The encoder processes the input sequence into a high-dimensional representation that preserves its temporal dependencies. The decoder then uses this contextualized representation to generate point or probabilistic forecasts. Figure 1 shows the vanilla transformer models which integrate the above concepts while addressing the computational limitations and improve the accuracy of the prediction tasks for energy-specific use cases.

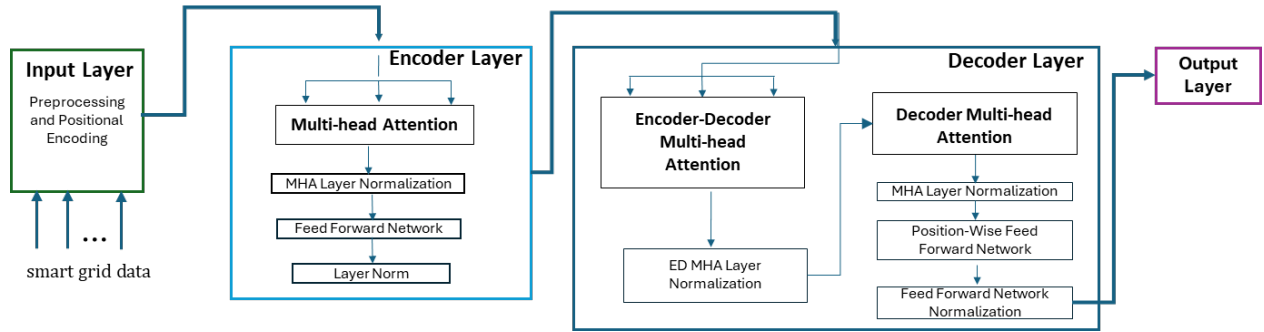

**Figure 1.** Vanilla transformer for predictive tasks in smart grids

Nowadays literature reviewed in section 4 has been built upon these foundational building blocks by introducing various architectural modifications to better address the challenges of energy forecasting and of vanilla transformer models [19-20]. Researchers have modified the attention mechanisms by implementing sparse or convolutional attention to efficiently handle long sequences or focus on important local changes [12], [21]. Energy systems often present spatial correlations, particularly in renewable energy generation patterns and in load consumption behaviors. Integrating graph attention mechanisms with transformers allows effective modeling of interdependencies among geographically distributed smart meters or generation units, boosting the forecasting precision at scale [22-23]. Other models introduced sparse and local attention mechanisms that were brought up to improve computational efficiency for long sequences [13], [21]. In this way, the tweaks that were added to the attention modules can enable the models to prioritize the more important intervals, such as peak load periods or sudden changes, achieving a lower computational overhead. Finally, output layers have been extended to produce not only point forecasts, but also probabilistic and multi-scale predictions, to offer a more reliable decision-making context in energy management [24-27].

The transformer-based models laid the foundation for LLMs, which are based on pretraining on immense volumes of heterogenous data, followed by task-specific finetuning. They represent a major paradigm shift in smart grid management having applications well beyond prediction tasks [28-29]. These models had a better generalization capacity than their predecessors, confirming that scaling up data volumes and model dimensions can lead to improved results [30-31]. Moreover, LLMs became effective even with small amounts of data due to their few-shot and zero-shot learning features and can be finetuned using multimodal inputs and heterogenous data. This holistic approach of fusing structured and unstructured data such as time series, weather dynamics, text documents and human behavior allows for better modeling of energy systems and decision-making support [32-33]. Therefore, LLMs are adaptable to a wide array of different smart grid management tasks. As depicted in Figure 2, the development of smart grid adapted LLMs involves a process composed from an initial pre-training step on broad knowledge, followed by fine-tuning on energy domain-specific data. This process facilitates the translation of general language understanding into specialized smart grid management intelligence. Fine-tuning does not mean retraining the model from scratch and can be done using smart grid logs, time-series from smart meters, reports, documents etc. Instead, it modifies only a subset of model parameters employing techniques such as Low-Rank Adaptation or prompt-based tuning to avoid overfitting.

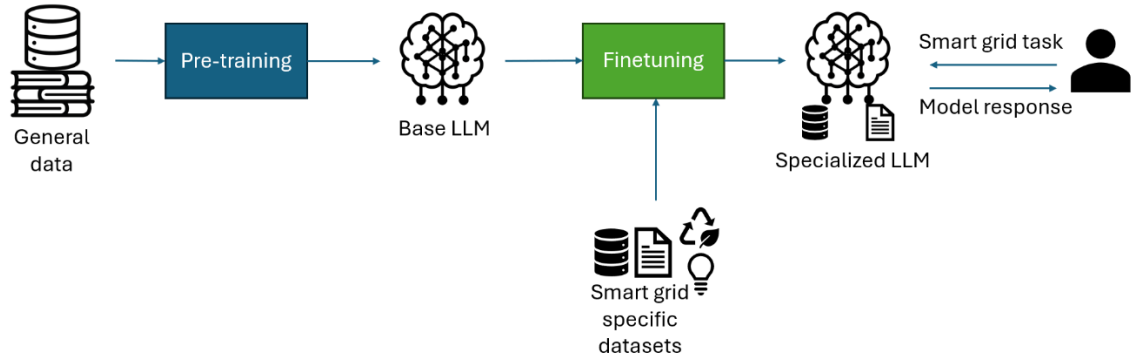

**Figure 2.** Finetuning process for LLM adaptation to smart grid management tasks

More specifically, according to the literature reviewed in this paper, LLMs have been applied for tasks related to industrial carbon emissions prediction, electrical safety incident detection, wind power and load forecasting and grid management by leveraging their pattern recognition, reasoning capabilities and generating decision-relevant information over complex multi-dimensional datasets. But more importantly, most of the studies focus on buildings and particularly on Building Energy Modelling. They are utilized to automate simulation tasks from data extraction to debugging, enabling accurate, scalable solutions for urban energy analysis. Furthermore, emerging workflows employ multi-agent (agentic) LLM workflows to fully automate complex tasks such as building energy modeling pipelines and retrofit recommendation systems. These developments not only streamline energy modeling processes but also pave the way for broader deployment of agent-based DT architecture for the energy sector, as highlighted in the perspectives and challenges section of the paper.

## 3. Methodology

The current systematic review aims to offer a comprehensive overview for the domain of applying transformer and large language models in the energy field focusing on identifying challenges and future directions. Therefore, we have used the Preferred Reporting Items for Systematic Reviews and Meta-analyzes (PRISMA) methodology that offers clear guidelines for conducting a systematic review and is widely accepted and recommended in literature for review articles [34]. The updated version of the PRISMA methodology from 2020 consists of a set of items to be included in systematic reviews and a flow diagram that sketches the process [35].

The first step in the PRISMA methodology is to provide an explicit statement of the research objectives the review addresses. In our study we have defined the following main objectives:

- Identify energy efficiency use cases and scenarios that are based on transformer models integration in smart energy grids
- Analyze the application of LLMs in energy systems management and optimization processes
- Investigate the challenges and determine the future research directions

The next step dealt with defining the main key phrases to be used in the literature search to cover the techniques, models and technologies to be analyzed in conjunction with the above three main research objectives. We have defined two categories for the search key phrases: transformer models in smart grid and LLMs in Energy Systems management and optimization. We then carry out separate searches combining the following keywords:

- Transformer / Large Language Model / Generative AI smart grid
- Transformer / Large Language Model / Generative AI energy / electricity grid
- Transformer / Large Language Model / Generative AI energy prediction / forecasting
- Transformer / Large Language Model / Generative AI energy efficiency / optimization
- Transformer / Large Language Model / Generative AI energy management

LLMs usage for energy management is a novel research direction with a lower number of approaches in literature. Thus, we have also used generative AI keywords to cover a larger number of articles. For our study, we selected Web of Science (WoS) [36] as the primary scientific database for conducting the search together with the Clarivate platform to perform WoS searches, applying the "Topic" criteria to identify relevant papers. The search and results processing activities have been done in February-April 2025.

Following the PRISMA 2000 sequential items, using the above objectives we have defined a set of inclusion and exclusion criteria for the review (see Table below).

**Table 1. Defined inclusion and exclusion criteria**

| Inclusion Criteria | <ul><li>Type of publication: article or proceeding</li><li>Article Language: English</li><li>Publication interval (last 5 years): 2021–2025</li><li>Top 4 publishers from WoS in the targeted research areas: Elsevier, IEEE, ACM and Wiley.</li><li>Included in computer science related categories</li></ul> |
|---|---|
| Exclusion Criteria | <ul><li>Duplicate (obtained as a result in different searches)</li><li>>= 10 citations for 2021 & 2022; >=5 citations for 2023; >= 3 citations for 2024; no citation constraint for 2025.</li><li>Review articles incorrectly marked as Article or Proceedings in WoS</li><li>Not properly covering the main research topic of transformer models/LLMs for smart energy grids</li><li>Duplicates (obtained as a result in different searches, kept only in one keyword category result)</li><li>Articles that could not be accessed as full text through the available institutional access</li></ul> |

Figure 3 presents the PRISMA 2020 flow diagram used in our study. It begins with the Identification phase, which includes compiling all results retrieved from the WoS database totaling 8550 papers. After removing duplicates, 7270 unique articles remained. In the screening phase we apply sequentially the inclusion and exclusion criteria, narrowing the number of articles to 1492 and finally to 99 included ones. A major complexity for the search was given by the meaning of the transformer word that refers mainly to the component that transfers electrical energy from one electrical circuit to another one and it is also used as term for the deep learning architecture. This issue spawned multiple results and implied difficult manual work in the screening phase. We have also tried to overcome this problem by combining transformers with model but still many results from the raw search referred to the transformer electrical component. At the same time LLM combination with energy efficiency key phrase gave as output many papers related to investigating the energy efficiency of LLMs execution, not with their application in the smart grid domain. similar to the transformer search, these papers have been filtered manually.

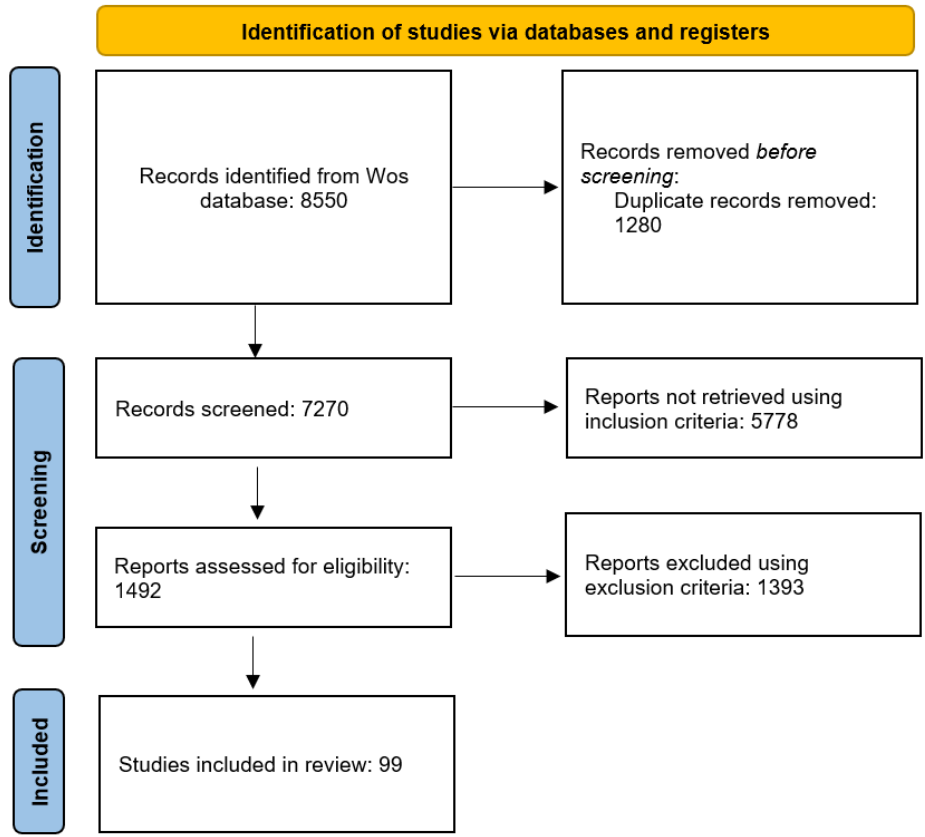

**Figure 3.** PRISMA 2000 flow diagram for the current study

The selected papers are grouped in Table 2 using defined search phrases and keywords. As can be noticed for some of the search phrases there were no articles included in the review. This is justified by removing many papers using strict exclusion criteria such as not properly covering the main research topic of transformer models/LLMs for smart energy grids or number of citations.

**Table 2. The selected articles distribution per search process**

| Keyword | Raw search result | Initial result based on inclusion | Result based on the exclusion criteria | ID |
|---|---|---|---|---|
| Transformer model energy prediction | 850 | 107 | 21 | [19-20], [27], [37-55] |
| Transformer model energy forecasting | 520 | 133 | 31 | [11-13], [21-26], [56-76] |
| Transformer model smart grid | 698 | 182 | 9 | [77-85] |
| Transformer model energy grid | 1381 | 149 | 9 | [86-94] |
| Transformer model electricity grid | 357 | 40 | 2 | [95-96] |
| Transformer model energy efficiency | 1403 | 216 | 2 | [97-98] |
| Transformer model energy optimization | 1078 | 145 | - | - |
| Transformer model energy management | 791 | 110 | 1 | [99] |
| Large Language Model energy prediction | 204 | 41 | 4 | [100-103] |
| Large Language Model energy forecasting | 86 | 11 | 4 | [28-29], [32], [104] |
| Large Language Model smart grid | 53 | 16 | 1 | [105] |
| Large Language Model energy grid | 97 | 24 | 2 | [30-31] |
| Large Language Model electricity grid | 17 | 5 | 0 | - |
| Large Language Model energy efficiency | 322 | 75 | 2 | [106-107] |
| Large Language Model energy optimization | 179 | 69 | 3 | [15], [108-109] |
| Large Language Model energy management | 169 | 50 | 4 | [110-113] |
| Generative AI energy prediction | 55 | 12 | 1 | [114] |
| Generative AI energy forecasting | 20 | 8 | 0 | - |
| Generative AI smart grid | 16 | 3 | 2 | [115-116] |
| Generative AI energy grid | 24 | 8 | 1 | [117] |
| Generative AI electricity grid | 7 | 3 | 0 | - |
| Generative AI energy efficiency | 88 | 34 | 0 | - |
| Generative AI energy optimization | 70 | 26 | 0 | - |
| Generative AI energy management | 65 | 25 | 0 | - |
| **TOTALS** | **8550** | **1492** | **99** | - |

Figure 4 details the mapping of the selected articles on publication year and their distribution per publisher. Most of the papers are published in the 2023-2025 timeframe, while as publisher Elsevier counts for 77% of the main four selected top publishers.

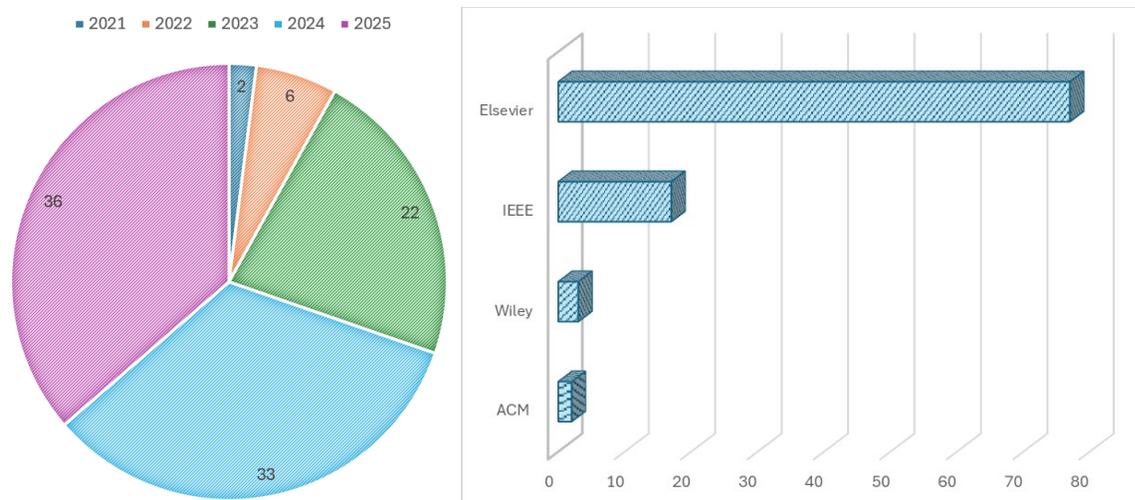

**Figure 4.** Paper distribution per publishing year and publisher

As for the accessing rights for the articles, Figure 5 shows the distribution of papers per access type highlighting that subscription-based access articles are more than double as the ones published under gold open access rights, but they can be still accessed through institutional subscription. Most of the included articles are published in journals, only five belonging to conference proceedings. Judging after the quartile of the included journals, 80 articles are indexed under quartile Q1 while 13 are Q2.

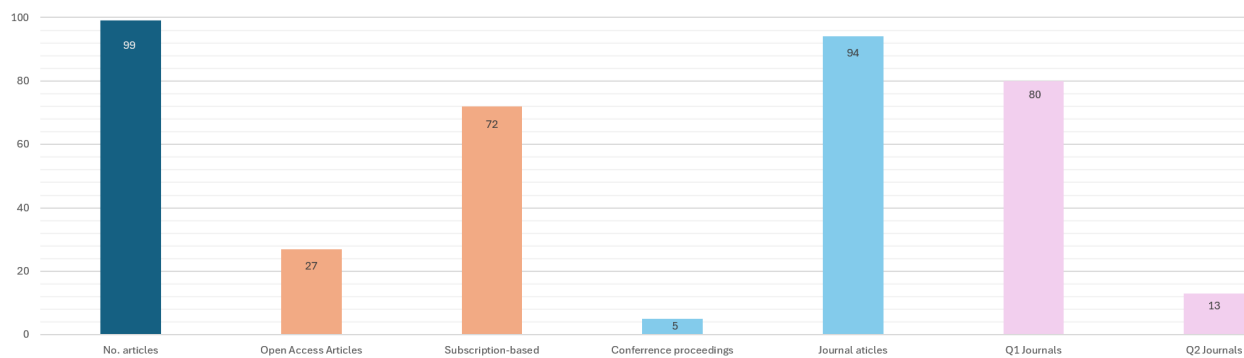

**Figure 5.** Access type, medium and quartile statistics

Considering the journals that published these articles, it can be noticed that there are four Elsevier journals that published more than five articles included in the study (see Figure 6): Applied Energy, Energy and Buildings, Renewable Energy and Energy. It seems that these journals tackle a lot the subjects investigated in our study. For IEEE most of the works are published in IEEE Access and IEEE Transactions on Instrumentation and Measurement.

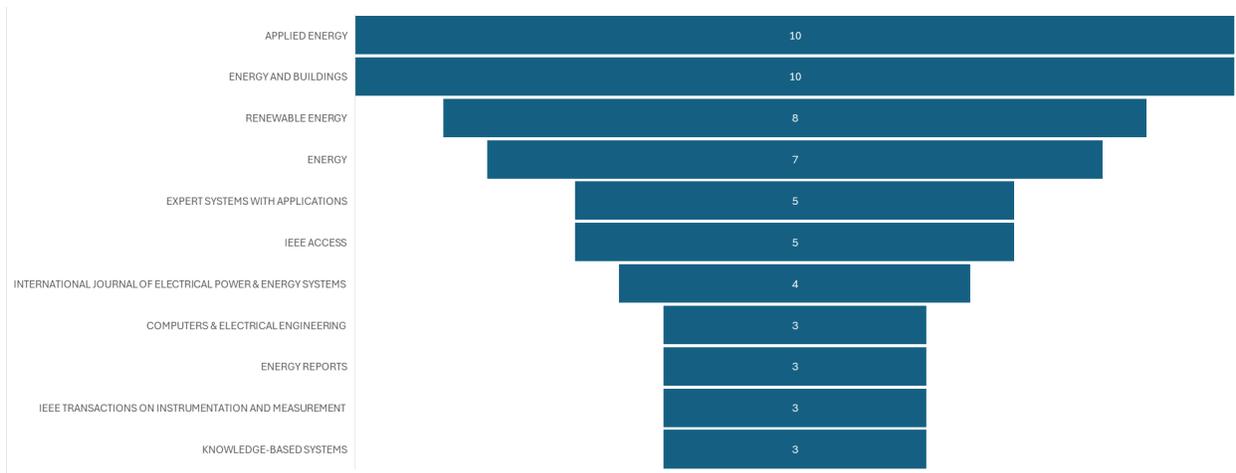

**Figure 6.** Number of articles per journal

The mean number of citations per paper is 16, showing a good impact on the included articles. However, more than half of these papers are published in 2024 and 2025 and some of their citations are not indexed. This also highlights the novelty of the targeted research domains. Considering distribution by author location, the figure below shows that more than 20 authors that tackle research problems related to transformers and large language models in the energy field are from Europe, more than 10 are from the US while this statistic is dominated by authors from Asia with more than 70 authors (see Figure 7). However, there are papers where co-authors come from different countries show that the research activity is a common objective for researchers from different parts of the world.

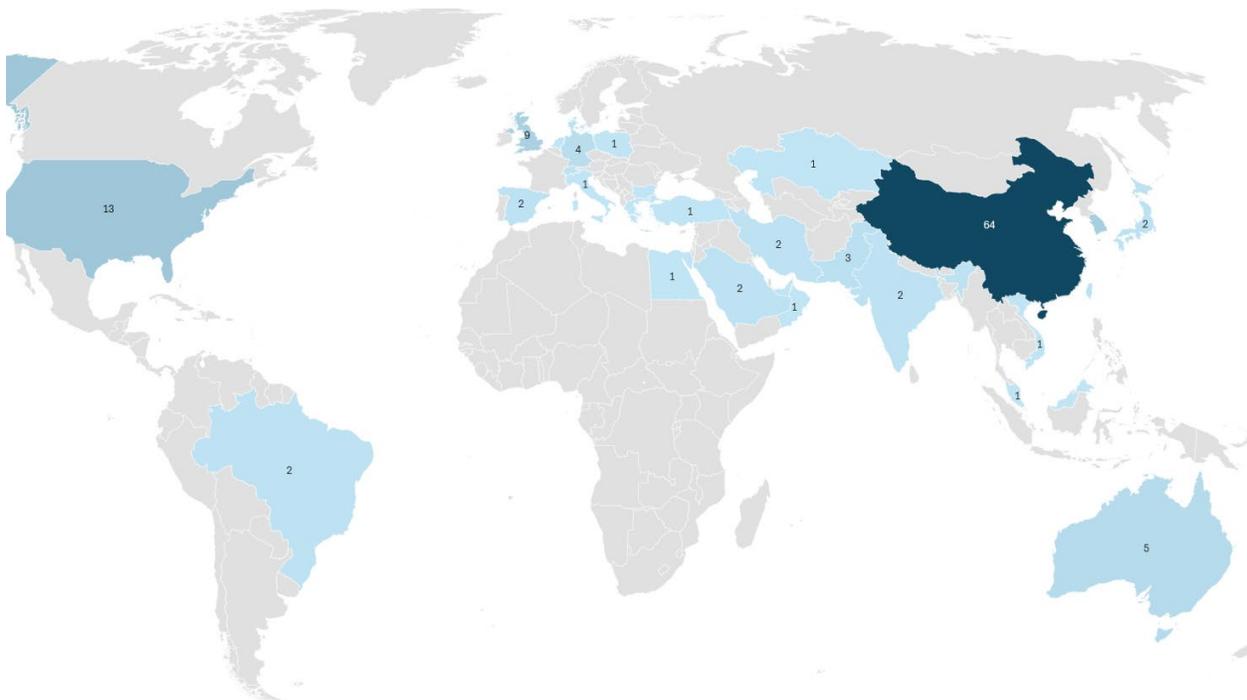

**Figure 7.** Country distribution of included articles

## 4. Transformers models in smart grids

In the context of smart grids, accurate forecasting of electricity and thermal demand and renewable generation is important for data driven decision-making to ensure efficient energy management, and grid reliability. Transformer models in smart grids are mostly designed to tackle energy forecasting tasks therefore we have organized the review to reflect the applications and challenges related to relevant components of a smart grid's energy flow

### 4.1. Electricity demand prediction

Transformer models are largely used for household-level electricity predictions aimed at control strategies for home batteries, and appliances. Yuan et al. [56] proposed a kernel-based real-time adaptive programing (ADP) framework for energy management systems, using a gated recurrent unit-bidirectional encoder representations from the transformer GRU-BERT. It fine-tunes BERT on time-series data and classifies the input loads by their type so that it can improve short-term household energy forecasting. This integration brought improved results over other baseline controllers. However, its complexity and extensive hyperparameter tuning phase are the downsides. Zhao et. Al [37] developed the STformer, a variation of the transformer architecture with spatial and temporal attention modules. This is used to yield better results on short-term household consumption forecasting tasks by learning the inter-residential correlations. Monte Carlo dropout enables probabilistic forecasting by quantifying uncertainty without requiring prior geographic information. Iqbal et al. [71] introduced an edge computing and transfer learning framework using an inverted transformer for real-time load forecasting. By inverting the transformer embedding, the model independently processes each variable, and it outputs forecasts with minimal historical data. While effective for new buildings, the model's success depends heavily on the quality of pre-training data and fine-tuning. For community-level energy forecasting spatial relationships between households, feeders, or neighborhoods are important. Zhao et al. [22] introduced a spatio-temporal graph attention transformer for multi-step residential load forecasting. The smart meter monitoring from multiple houses is modeled as graph data, with edges representing physical or functional similarities. They design spatio-temporal graph attention blocks within a transformer encoder-decoder architecture to capture inter-household correlations like voltage drop effects or synchronized usage patterns, and temporal self-attention for each house's time series. This model significantly improved prediction accuracy but faces challenges with keeping graph topology updated and demands higher computational resources. Dong et al. [73] developed a deep transformer seq2seq model to forecast household-level reactive power demand in short-term, as an extension to active power forecasting. The model's capacity to integrate multiple inputs like voltage and temperature improved the forecast accuracy. Also, Shan et al. [85] introduces a Multiscale Self-Attention Network (MSANet) for nonintrusive load monitoring (NILM), aiming to disaggregate total household energy consumption into appliance-level loads. A dilated window self-attention mechanism and a multibranch structure are employed to capture both global temporal correlations and local sequential features, while also addressing data imbalance issues through specialized subtask networks. The self-attention-based model outperforms baselines, achieving superior MAE and F1 scores in energy disaggregation tasks.

Common open challenges include transformer model complexity and interpretability. Xu and Chen [12] introduced an interpretable transformer model for probabilistic residential net load forecasting, which employs a sparse self-attention mechanism and an embedded feature selection network to highlight the most influential inputs. Similarly, Santos et al. [24] demonstrated that transfer learning can mitigate data scarcity in building load prediction by pre-training a temporal fusion transformer on data-rich buildings and fine-tuning it for new buildings with low data volumes. The method achieved lower errors in local energy communities. These approaches aim to reduce the black-box nature of common AI models and high data requirements of transformers in the residential context. Wang et al. [26] propose PDFformer, for customer electricity demand forecasting. They integrate a probabilistic decoder and an adaptive online probabilistic forecasting mechanism (AOPF), so that the model can dynamically update only the probabilistic decoder based on real-time loss evaluation, improving computational efficiency while maintaining high forecasting accuracy. Semmelmann et al. [68] addresses forecasting in communities with rising heat pump presence showing that Transformer-based models outperform tree-based and traditional recurrent methods, particularly when forecasting aggregated loads directly. In addition, while the proposed ensemble empirical mode decomposition with adaptive noise improves the performance of Transformer and LSTM, it negatively impacts tree-based methods. A sparse attention-based transformer for short-term load forecasting is proposed to address inefficiencies in LSTMs based predictions [21]. Their encoder-decoder model reduces computational cost while maintaining accuracy. Furthermore, Abdel-Basset et al. [78], introduced Energy-Net a deep learning model for short-term energy consumption forecasting in IoT-enabled smart grids. The model stacks multiple spatiotemporal modules, each comprising a Temporal Transformer (TT) for capturing temporal load patterns and a Spatial Transformer (ST) with improved self-attention and convolutional layers to extract spatial features. Notably, the model maintained low computational complexity suitable for resource-constrained IoT devices connected to smart grids. Giacomazzi et al. [65] used household-level data but focused forecasting at the substation and grid levels. They showed that training TFT models on substations and then aggregating the forecasts yielded better accuracy than direct grid-level prediction. This hierarchical approach allowed the model to learn load patterns and improve aggregate forecasting accuracy. To achieve a balance between speed and accuracy MEAformer architecture is proposed [19]. The self-attention is replaced with a linear external attention mechanism using an all-MLP decoder. This lightweight design helps in reducing the computational demands while maintaining strong forecasting accuracy, particularly for household-level electricity consumption. Also, Nazir et al. [79], presented a daily, weekly, and monthly load energy consumption prediction model using the TFT. By leveraging both primary and auxiliary data sources and using batch training techniques, the TFT model outperforms traditional deep learning models. Rong et al. [45] introduced a transfer learning-enhanced transformer for improved accuracy of household appliance identification. The transformer is pre-trained on one set of houses and fine-tuning on another to address data scarcity and generalization issues. The transformer's multi-head attention was better at isolating the patterns of each appliance from the aggregated data, especially after fine-tuning to a new home's specifics. This model outperformed baseline models in both single-load and multi-load scenarios, obtaining lower errors on multiple datasets.

Several studies address the limitations of traditional transformers for long-term energy forecasting. Wang et al. [20] propose PWDformer, which improves temporal learning by integrating deformable-local aggregation, position weights, and a frequency selection module, leading to accuracy gains across benchmark datasets. Similarly, Su et al. [48] introduced MDCNet, that improves long-term forecasting at household level by transforming 1D time series into 2D feature spaces using mode decomposition and 2D convolutions. The model first applies variational mode decomposition to split input series into a trend and multiple intrinsic mode functions across frequency bands, then processes them with specialized prediction blocks and convolutional modules to better learn the correlations between intramodal and intermodal features. Hierarchical transformers that generate initial probabilistic forecasts and then refine them via a conditional generative model are proposed in [27]. The time series decomposition into trend and seasonal components and correction of the transformer's outputs effectively reduces errors over long forecasting horizons. Huang et al. [43] propose CrossWaveNet, a dual-channel network for household-level electricity forecasting. The model decomposes the input signal into seasonal and trend-cyclical components, processing them separately through LSTM/GRU-based modules before merging for the final prediction. They additionally propose FL-Net [46] for household-level electricity consumption forecasting, combining reversible instance normalization, multi-scale decomposition, and frequency external attention. Their model separates trend and seasonal components and then refines them through cross-decomposition, achieving strong long-horizon forecasting performance at a lower computational cost. Similarly, Saeed et al. [23] propose the SmartFormer, a graph-based transformer household and regional smart grid electricity load forecasting. The model improves forecast accuracy across multiple datasets, addressing challenges of inter-series dependencies in multivariate smart grid data. Yu et al. [53] developed Robformer, which embeds a second-order differencing decomposition and a specialized trend forecasting block into the transformer to better handle abrupt load shifts and irregular seasonal fluctuations. This improves the model's stability and reduces errors when system behavior changes suddenly. Nonetheless, fully addressing regime changes may require techniques like continual online learning or periodic fine-tuning to keep models updated. An interesting approach for secure load management in smart cities is proposed by Wang et al. [98]. An AI-driven Multi-Stage Learning-to-Learning (MSLL), leveraging MMStransformer which is tailored for multivariate data and long-rage dependency capture in load forecasting. Computational efficiency is optimized while high prediction accuracy is maintained, while it integrates environmental and operational variables to handle the complexity in smart city data.

Forecasting aggregated loads supports long-term decisions on grid upgrades and substation expansions. Huy et al. [64] applied the Temporal Fusion Transformer to city-level load forecasting in Hanoi, Vietnam, obtaining superior performance over traditional statistical models and RNNs. TFT's architecture combines LSTM-based local processing with self-attention, with the model learning both short and long-term load patterns helped by the integration of weather and calendar features. In the Hanoi case, TFT reduced day-ahead forecast errors compared to LSTM and ARIMA models. Cao et al. [91] proposes a Shifted Window Attention Transformer model for power distribution prediction, designed to effectively integrate both global trends and local variations in time series data. Self-attention within varying window sizes and the application of patch relative position encoding makes the model capture spatiotemporal patterns more accurately and operate in a bottom-up fashion to progressively extract features from local to global levels. Cui et al. [61] address provincial-level electricity load forecasting by augmenting the Informer model with a Season-aware Block designed to learn the low-frequency seasonal patterns and temporal covariates. Their dual-path architecture models both high-frequency variations and long-term seasonal trends, improving long-horizon forecasting accuracy for large-scale power systems. Similarly, Du et al. [67] propose MOWOA-Transformer for regional electricity load forecasting from New South Wales, Australia, by combining a multi-objective whale optimization algorithm (MOWOA) with a transformer model to automate hyperparameter tuning. Their optimized model achieves up to 35% MAPE reduction compared to unoptimized baselines and provides accurate probabilistic forecasts, addressing the dual needs for precise load prediction and uncertainty estimation in large-scale power systems. Gao et al. [58] propose Adaptive-TgDLF, a transformer-based model for district-level electricity load forecasting that integrates domain knowledge into the training process. By decomposing load into a smooth trend and local fluctuations, and using physics-guided methods for the trend, the model performs well even when anomalies and noise are present. Additionally, Ferreira et al. [63] showed that interpretable transformer models can achieve high accuracy in substation-level load forecasting. They used the temporal fusion transformer with built-in variable selection and attention mechanisms, to forecast aggregated substation loads up to 48 hours ahead with a MAPE below 1.5%. Qingyong Zhang et al. [54] also address the national-level net load forecasting task aiming to improve grid stability by learning multi-scale temporal patterns and inter-variable dependencies. The transformer model is augmented with temporal-channel attention block, combining temporal and channel attention within a hierarchical encoder–decoder using multi-scale patch-based embeddings. Tested on country-wide datasets from Austria, Belgium, and France, it consistently outperforms state-of-the-art models, though its computational demands and hyperparameter sensitivity may limit deployment in real-time scenarios. Similarly, Zhang et al. [69] propose TransformGraph, a hybrid model for national-level electricity net load forecasting that combines a transformer with a graph convolutional network (GCN). The model performs well, learning feature correlations via LSTM-derived graphs and the temporal dynamics within the data. It results in improving short-term forecasting accuracy on OPSD datasets from Austria, Belgium, and Slovenia, outperforming ARIMA, RNN, LSTM, and vanilla transformer models. Finally, Zheng et al. [72] introduced a framework that integrates spectral clustering-based decomposition with temporal fusion transformers to forecast aggregated building energy use from different building zones (offices, dormitories). Unlike conventional deep learning methods that rely on raw energy data, their proposed approach first applies spectral clustering to decompose building energy consumption into multiple interpretable subsequences, which are then encoded as vector representations and used as static covariates in a temporal fusion transformer model. Despite its advantages, the reliance on high-resolution sensor and weather data, along with the complexity of spectral clustering, may limit its adaptability across different building types.

Finally, transformers have also been applied to specialized forecasts like EV charging demand. Gao et al. [76] used a multi-attention fusion transformer to predict aggregate residential load under coordinated EV charging. They use a pre-trained temporal fusion transformer with minimal EV-specific data improving the forecast accuracy compared to baseline models like MLP and LSTM, showing that pre-trained transformers can perform well in data-scarce environments. Dong et al. [41] applied transformer models to improve the prediction of energy consumption for electric buses. Their approach uses a transformer encoder to extract implicit features from driving behavior and vehicle dynamics data, which are then fed into a decoder for energy consumption forecasting. By learning the relationships between driver behavior and real-time bus driving, the transformer achieved more accurate consumption estimates across varying driving distances. Furthermore, Jiang et al. [96] developed a V2X (vehicle-to-everything) value-stacking framework that coordinates vehicle-to-building (V2B), vehicle-to-grid (V2G), and energy trading to maximize economic benefits for residential communities while maintaining grid stability. It employs a dynamic rolling-horizon optimization (RHO) method and introduces a Transformer-based forecasting model—GRU-EN-TFD (Gated Recurrent Units-Encoder-Temporal Fusion Decoder)—to handle uncertainties in building loads, solar generation, and EV arrivals. Simulation results demonstrated that the Transformer-based model outperformed all benchmarks.

In [81], Ma et al. propose a muti-branch cross-Transformer integrated into a prior-guided state-reused network that aims to non-intrusively perform load disaggregation. The Transformer block is the central component and handles the feature fusion from multiple sources, including an appliance operating power pattern to perform the disaggregation. As seen, many electricity demand prediction transformer models involve elaborate architectures that may be difficult to deploy at scale. There is ongoing work to simplify these or apply transfer learning so that one transformer model may address multiple energy assets of different energy scales with minimal retraining or computational resources and in scarce data scenarios. Despite these challenges, transformer models have clearly demonstrated superior accuracy in electricity demand forecasting, achieving lower errors and enabling better control decisions than previously possible. However, LLMs offer promising solutions to these challenges due to their relevant features of transfer learning and time series energy data augmentation. Table 3 presents a comparative view of the research done for electricity demand prediction using transformer models.

**Table 3. Analysis of the transformer models for electricity demand prediction**

| Encoder-Decoder | Attention Mechanism | Architectural Extensions | Work |
|---|---|---|---|
| **Household/Residential Load +/- Weather data** | | | |
| BERT encoder; GRU decoder | Standard scaled dot-product | Kernel-based ADP | [56] |
| LSTM-based Transformer | | Multi-task learning, variable selection module | [73] |
| iTx blocks with self-attention, FFN, layer norm | | N/A | [71] |
| TFT architecture | | | [79] |
| 2 encoder layers with autocorrelation, spatial attention, and decomposition | Autocorrelation-based multi-head mechanism | MC Dropout for probabilistic forecasting | [37] |
| STGA encoder and decoder | Spatial, temporal, and transform attention | Gated fusion for combining spatial/temporal patterns; graph-based Seq2Seq for inter-household modeling | [22] |
| No encoder-decoder, direct generative Transformer | Sparse multi-head self-attention | Local variable selection module | [12] |
| Static encoder for building context + LSTM for local patterns | Temporal self-attention | Variable selection; gating mechanisms | [24] |
| MLP encoder with temporal external attention + MLP decoder | Temporal external attention across sequences | All-MLP structure, shared external memory, reversible normalization | [19] |
| Multibranch cross-Transformer | Cross Attention mechanism | Convolutional Fusion Block and Multi-head Self attention | [81] |
| **Aggregated residential load (building level) +/- weather data** | | | |
| Probabilistic decoder | Standard scaled dot-product | Conv1D input layer; AOPF module | [26] |
| TFT architecture | | DECPR spectral decomposition | [72] |
| CNN + pooling for local patterns | | N/A | [45] |
| Standard | | CEEMDAN decomposition per IMF | [68] |
| | Standard scaled dot-product + deformable local aggregation | | [20] |
| | Event-based + ProbSparse attention | CSPBlock for lightweight training, TSDBlock for seasonal/trend decomposition | [76] |
| LSTM encoder | Temporal self-attention | Hierarchical aggregation from substations | [65] |
| **Household/Residential & Aggregated Load +/- Weather data** | | | |
| Encoder with GNN | ProbSparse attention for long-sequence efficiency | Graph structure learning | [23] |
| Conditional generative decoder for hierarchical forecasting | Standard scaled dot-product | Hierarchical probabilistic forecasting | [27] |
| Dual LSTM/GRU-based channels for seasonal and trend features | Cross-channel dot-product | Dual-channel architecture | [43] |
| Encoder only | Frequency external attention | Multi-scale trend-season decomposition, external memory | [46] |
| Trend and modal streams via MLP and 2D CNNs | No attention module | Variational Mode Decomposition (VMD) | [48] |
| Standard | Auto-correlation mechanism with lower complexity | Trend forecasting block, decomposition block | [53] |
| Dilated window transformer blocks and distilling layers | Multi-scale self-attention mechanism | N/A | [85] |
| **Aggregated district/city/grid Load +/- Weather data** | | | |
| Standard | Standard scaled dot-product | MOWOA for hyperparameter tuning, longitudinal data selection | [67] |
| | | Domain-informed decomposition, transfer learning | [58] |
| | | Deep feature extraction module from dynamic bus states | [41] |
| | | GCN-based feature aggregation, LSTM-derived graph structure | [69] |
| TFT architecture | | Value stacking rolling horizon optimization problem | [96] |
| | | Variable selection networks, gated residual layers | [63] |
| Multi-scale standard encoder-decoder structure | Temporal-channel attention | Multi-scale patch embedding | [54] |
| Informer encoder-decoder | ProbSparse self-attention | Parallel seasonality MLP block for artificial covariates | [61] |
| LSTM encoder | Temporal self-attention | Linear regression pre-trend removal, gated residual blocks | [64] |
| TT and ST architectures | Spatial attention mechanism | Stacked spatiotemporal modules | [78] |
| Vanilla transformer | Multi-head window attention | Shifted window attention blocks | [91] |
| Input embedding and transformer blocks. | Multi-head self-attention mechanism | Multi-mask learning-to-learning strategy | [98] |

## 4.2. Thermal energy demand prediction

Transformer models have been also used to forecast multi-energy demands. Wang et al. [57] developed a multi-energy transformer for a campus energy system that integrates electricity, cooling, and heating loads. Their model, MultiDeT (Multi-decoder Transformer), uses a single transformer encoder to process the historical profiles of all energy loads together, and then uses separate transformer decoders for each load type's forecast. The encoder's self-attention effectively learns the shared patterns and correlations between the different energy datasets. Meanwhile, each decoder can focus on the particularities of one load. This architecture obtained better results than forecasting each load with an independent model. The model obtained lower errors, alongside faster training convergence since the shared encoder learns a common representation for all tasks. Zhang et al. [25] introduced a probabilistic Patch Time-Series Transformer for integrated load forecasting. This model segments each energy load time series into patches or subsequences and feeds them into a transformer, like how vision transformers handle image patches. It then applies quantile regression at the output to estimate prediction intervals for each energy type, achieving more accurate forecasts for building electricity, cooling, and heating loads. Yan et al. [59] proposed an improved feature-time transformer with a Bi-LSTM decoder (FTTrans-E-BL) for user-level integrated energy forecasts. The most important result was that the model outperformed multiple benchmarks across multiple load types simultaneously. By visualizing attention patterns, they showed that the transformer learned meaningful correlations. Similarly, Cen et al. [62] developed a model for short-term multi-load energy forecasting in buildings that uses a channel-independent forecasting strategy combined with a patching mechanism and a temporal convolutional network followed by a stacked transformer encoder. However, while the model performs good in independent load forecasting (air conditioning load forecasting), it does not explicitly account for inter-channel dependencies.

Recent studies show that transformer architecture can improve building load predictions by learning complex occupancy and thermal patterns from data. Li et al. [40] proposed a transformer-based encoder–decoder model for short-term cooling load forecasting in large commercial buildings. Their model encodes past cooling load and influencing factors such as weather data into latent features via multi-head attention, then decodes the output to predict future cooling demand. After training, they computed attention-based importance scores for each input feature to rank their influence on the cooling load forecast. This provided valuable interpretability, confirming for instance that outdoor temperature and solar irradiance were the top prediction influences. Faiz et al. [75] compares a Transformer-based model and a CNN-LSTM hybrid model across multiple forecasting horizons showing that the Transformer model consistently outperforms the CNN-LSTM. Also, they provide an analysis of how forecasting aggregated indoor temperatures can impact energy optimization and occupant comfort. Wang et al. [47] introduced a transformer variant that decomposes a building's energy (heating and cooling stations, lighting, elevator, etc.) data into distinct trend and periodic components and applies a specialized spectral-patch attention mechanism. In evaluations on a large office building, SPAformer achieved 12% lower error than vanilla transformer models, demonstrating the benefit of explicitly attending to multiple different patterns in the data. In another study, Die Yu et al. [55] developed a hybrid approach for a hub airport's cooling load that integrates clustering with transformer prediction. They first grouped daily cooling load profiles by using an improved weighted-DTW K-means algorithm, then fed cluster features and labels into a temporal fusion transformer. This approach effectively learned the influence of an airport's operational schedule and thermal energy storage on cooling demand.

Forecasting short-term natural gas demand is an important piece for operational planning, yet challenging due to irregular patterns and weather sensitivity. Han et al. [50] proposes a production prediction model for liquefied petroleum gas (LPG) plants that integrates the Boruta algorithm with a CNN-based transformer model. The model uses CNN for adaptive feature extraction and a transformer layer to learn the temporal dependencies across different representation spaces. This integrated approach outperforms traditional models in terms of prediction accuracy, offering practical operational guidance. Lin et al. [70] introduced time-enhanced perception transformer model employs a standard transformer encoder augmented with convolutional self-attention and enriched positional encodings for gas demand prediction. It is paired with a two-stage feature extraction step that decomposes the gas load time series into trend, seasonal, and residual components and generates informative lags. Using sequence modeling with domain-specific feature engineering, the model achieved high accuracy. Extensive tuning of the convolution kernel sizes, and feature lags were required for peak performance, therefore model's complexity could pose problems for real-time deployment in resource-constrained settings.

Finally, it's worth mentioning that transformer models are enabling cross-domain energy forecasts that account for interactions between systems. For instance, recent research has demonstrated the effectiveness of transformer-based approaches in simultaneously forecasting multiple variables, such as energy demand and market prices. Traditional forecasting models typically treat these variables in isolation, thereby overlooking the interdependencies that may or may not exist between them. In contrast, transformer models integrate information from multiple domains, thereby offering a more holistic view of the energy landscape. By using the concept of attention, transformers can assign appropriate weights to different features and quantify the impact of one variable on another.

**Table 4. Transformer based approaches for thermal energy demand prediction**

| Input Representation | Encoder-Decoder | Attention Mechanism | Architectural Extensions | Article |
|---|---|---|---|---|
| City-level gas load; temperature data | FFN decoder | Convolutional-based attention | STL & TLAC two-stage decomposition | [70] |
| Campus-level heating and cooling loads; weather data | One shared encoder and 3 load-specific decoders. PatchTST encoder | Standard scaled dot-product | Uncertainty estimation; patch-based time-series segmentation | [57], [25] |
| Building-level electricity, heating and cooling loads; weather data | TFT encoder & Bi-LSTM decoder. MLP block for long-term trends | Feature and timewise dual multi-head attention. Spectra-Patch Attention | WTLCC-based input lag selection. Trend-period decomposition | [40], [47], [59] |
| Floor-level air conditioning data; indoor conditions | PatchTCN encoder (encoder-only approach) | Standard scaled dot-product on sub-series | TCN for temporal embedding, patch segmentation for short-time patterns | [62] |
| Room-level HVAC temperature; indoor comfort survey | Encoder-only approach | Standard scaled dot-product | - | [75] |
| Airport-level cooling load and passenger flow; weather data | TFT architecture | Standard scaled dot-product | k-means clustering module of daily load patterns | [55] |

| LPG production thermal data | Standard with CNN layer before the encoder | Standard scaled dot-product | Boruta for feature selection | [50] |

### 4.3. Renewable generation prediction

Wind and solar power generation can be quite challenging to forecast due to the inherent variability of renewable energy sources and high dependence on accurate weather forecasts.

Transformer models have shown promise for wind power forecasting by learning short-term gust patterns and long-term seasonal effects. Early applications of transformers showed improvements over recurrent networks, and recent innovations target the wind's unique challenges such as extreme fluctuations and spatial correlations among turbines. Nascimento et al. [11] proposed a transformer-based model with integrated wavelet transforms to separate high- and low-frequency components of wind speed series. Their model, using cyclical positional encoding for daily periodicity, achieved better prediction accuracy than LSTM, GRU, and vanilla transformers, especially in extreme wind speed conditions. Liu and Fu [13] introduced the Graph Patch Informer (GPI), a self-supervised learning-enhanced transformer model designed for forecasting multiple renewable energy sources, including photovoltaic and wind power. The innovation consists in the integration of patch-based self-attention, graph attention networks, and a self-supervised pre-training strategy. GPI segments the time series into patches to preserve local continuity, then the graph attention network is employed with a self-adaptive adjacency matrix. Experimental results demonstrate significant improvements, compared to models like Autoformer across different renewable energy forecasting tasks.

Efficient handling of spatio-temporal relationships in wind farms can significantly improve transformer models' predictions. Wind speeds at neighboring turbines or locations are correlated therefore using spatial data can improve predictions. Therefore, transformer model variants incorporate graph attention, like the residential load case, or encoder-decoder models that use numerical weather predictions as additional sequences. Zefeng et al. [29] had fine-tuned a BERT model specifically for wind power forecasting (BERT4ST) and included spatial features like wind speed from adjacent sites. It is one of first approaches that aimed to adapt a language model to wind forecasting and consistently outperformed all analyzed baseline models, including variants like Informer and Autoformer, on multiple wind farm datasets. Its architecture successfully learned spatial correlations by treating them as another sentence in the input, and the self-attention layers then picked up on both temporal patterns and cross-site dependencies. The model was pre-trained on sequence data and fine-tuned on wind data with minimal architecture changes, beating specialized models, indicating the potential power of transfer learning in use cases wind farms. Similarly, a pre-trained language model with spatial prompts and trend/seasonal decomposition are combined to forecast wind speeds [28]. STELLM decomposed wind time series into trend/seasonal components via multi-scale pooling, tokenized the series into patch sequences, and then added spatio-temporal prompts to these token sequences before feeding into a large pre-trained model. These types of approaches address multiple challenges at once, such as data heterogeneity, long-sequence modeling, and extreme event forecasting, at the cost of a very complex and expensive computational model. Zhou et al. [44] focused mainly on wind's volatility defining a transformer model named Diformer that modifies the attention mechanism to better focus on changes in the energy time series, rather than on the absolute values. It computes bidirectional differences in the input sequence and applies self-attention to these difference series, making the model highly sensitive to sudden trend reversals, spikes or drops in the dataset. Additionally, Diformer uses a dynamic trade-off loss function that penalizes large errors more heavily outperforming other transformer-based models, in both short-term (15-min ahead) and ultra-short-term (5-min ahead) wind forecasting tasks.

One remaining challenge is the model scaling to longer horizons. Difference-focused attention performs well at very short horizons but might need additional components for day-ahead horizons. Conversely, Mo et al. [51] developed Powerformer, a transformer for wind power forecasting that integrates an LSTM-based embedding layer, sparse self-attention, and temporal pooling. They filter out temporal noise and improve feature extraction for wind data with the LSTM embeddings significantly reducing the errors compared to a basic transformer and to LSTM/GRU models, while also using less computation, thanks to the sparse attention. Wind power forecasting is closely linked to transformer-based methods for predicting wind speed, an exogenous variable that is a driver of power production. Wu et al. [52] uses transformer and graph neural network techniques for wind prediction. A transformer-based wind predictor is deployed at each production site and a multi-dimensional spatial-temporal GNN (MST-GNN) aggregates the information across neighboring sites. This combined approach improved multi-step wind speed forecasting accuracy by roughly 9% compared to models without spatial coupling. Wang et al. [38] propose a convolutional transformer-based truncated gaussian density network (TGDN) with a probabilistic forecasting approach. The model incorporates wavelet soft threshold denoising to filter noise from wind speed data, followed by maximal information coefficient-based feature selection to improve the quality of input representation. The hybrid architecture integrates convolutions for multi-scale spatial feature extraction and multi-head self-attention transformers for long-term temporal dependencies. Experimental results across three real-world datasets prove that this model achieves superior forecasting accuracy compared to LSTM, GRU, and standard Transformer models. Notably, the integration of wavelet denoising leads to a marked improvement over raw-data-based models, and the truncated Gaussian approach provides better uncertainty quantification compared to traditional quantile regression.

In addition to the purely data-driven approaches, we have identified a direction that explores hybrid transformers with physical-informed components. For example, some works merge transformer forecasts with numerical weather prediction outputs or use physics-guided augmentation to inform the attention mechanism [58]. This hybridization can improve performance in scenarios where historical data may not contain sufficient information for future wind changes. Liang et al. [88], ARFEAT, a robust short-term wind power forecasting framework is presented. The model architecture contained anomaly repair, feature enhancement, asymmetric loss optimization, and a transformer-based model. It uses self-attention to model temporal dependencies and an asymmetric loss to penalize overestimations, addressing energy trading risks. One of their biggest contributions is the inclusion of a semi-supervised anomaly detection (KNN-LightGBM) module and Fisher score-based feature selection. ARFEAT achieves good accuracy and cost reduction on datasets from inland and coastal wind farms. The work by Thiyagarajan et al. [92] introduces MST-Net, a modified twin transformer model for one-hour-ahead wind power forecasting. It combines GoogleNet feature extraction with Swin transformer blocks shifted window attention to learn local and global temporal patterns. It achieves good accuracy on real-time wind data from Tamil Nadu, outperforming baselines. Mirza et al. [93] present QT-MARF, a quantile-transformed multi-attention residual framework for medium-term wind and PV power forecasting. Their model is built on a transformer encoder-decoder architecture and being a quantile transformer, it normalizes inputs for improved accuracy and generalization. Trained on datasets from Brazil and China, QT-MARF outperforms models like CNN-GRU and CNN-LSTM, achieving higher $R^2$ scores.

Table 5: Comparison of Transformer-Based Models for Wind Power Forecasting

| Input Representation | Encoder-Decoder | Attention Mechanism | Architectural Extensions | Work |
|---|---|---|---|---|
| Wind speed + aggregated wind farm data + weather data | Informer encoder + linear decoder | Probabilistic attention with patch segmentation | Segment-wise patching; GAT for supervised learning | [13] |
| Aggregated wind farm data +/- weather data | Pre-trained BERT encoder (4-layer truncated) | Bidirectional self-attention | Spatio-temporal patch encoding; | [29] |
| | Swin encoder | Standard scaled dot-product with windowing | Siamese Transformer | [92] |
| | Standard encoder-decoder | Self-differential attention | N/A | [44] |
| | LSTM-based encoder | Sparse self-attention | | [51] |
| | 3 encoder layers | | Anomaly detection; multi-feature imputation | [88] |
| | Encoder-only approach | | Wavelet decomposition module | [11] |
| Aggregated wind speed data +/- weather data | Standard encoder-decoder | Standard scaled dot-product | Wind-Transformer at each node; GNN for spatial correlations | [52] |
| | CNN-based encoder, no decoder | | Wavelet soft threshold denoising, feature selection module | [38] |
| | Encoder-decoder with gated residual networks | | Quantile transformation module | [93] |

Transformer architectures have been applied to solar power forecasting to address similar challenges as in wind case, such as temporal variability, multivariate inputs and long-term dependencies. Zhang et al. [42] developed a multimodal model that combines ground-based fisheye sky images with meteorological data, using a gate-enabled transformer mechanism. In this approach, CNNs first extract visual features from sky images, while a transformer processes the time-series of recent irradiance and weather measurements. A gating network then combines the two outputs by weighing the CNN and transformer features, effectively deciding how much the current cloud cover should influence the forecast and how much the historical irradiance trend should influence it. The results highlight the strength of transformers in multi-modal data integration, by attending to different inputs (here, numerical and image data) through gating. Liu et al. [74] combines historical solar irradiance data with ground-based sky image features. An Informer model encodes historical and clear-sky global horizontal irradiance, and a vision transformer processes optical flow maps derived from sky images. A cross-modality attention mechanism fuses these features before a generative decoder produces multi-step forecasts. However, complexity and computational cost represent a challenge real-time deployment. Zhang et al. [39] proposes a hybrid Transformer-based framework for solar irradiance forecasting that is specifically designed to address the challenges posed by incomplete data and complex irradiance patterns. The framework integrates a mask-transformer data imputation module with a prediction module to accurately impute missing values by capturing contextual information to reduce noise in the input data. Concurrently, it makes use of series decomposition and feature fusion to extract and integrate patterns from the irradiance time series. Alorf et al. [66] proposes an N-hour-ahead forecasting model that first applies variational mode decomposition to decompose irradiance signals into mode functions, eliminating noise and handling non-stationarity. These components are then processed by a modified temporal fusion transformer that is coupled with a variable screening network and a GRU-based encoder–decoder. Yuan et al. [49] proposed a model with a multi-timescale fluctuation aggregation (MFA) attention mechanism, which segments PV power sequences into sub-series of varying lengths to separately learn short-term and long-term patterns. They also employ contrastive learning to identify similar historical days as additional context for the current time step. Zhang et al. address in [83] the problem of capturing the patterns in cross-batch time-series data for solar irradiance forecasting. The proposed Transformer model (CRAformer) uses Dual Cross Residual Attention mechanism and a dual output for the encoder. An additional Convolutional Weighted Fusion Module is integrated to improve feature extraction. The performance of CRAformer was evaluated on different prediction time steps and compared with the oTransformer model it performed better. Yin et al. propose in [84] a Transformer deep learning model for multi-feature timeseries prediction, as the data-driven component of a four-area integrated energy system. The performance of the Transformer was demonstrated on a PV dataset where it performed better than a Bi-LSTM model. In this case, the forecasted time series are fed into the reinforcement learning component to help the policy selection mechanism.

Despite the recent improvements in RES energy prediction brought by transformer models the open challenges of addressing multivariate inputs and long-term dependencies remain open, one promising solution being the integration of LLMs in broader pipelines augmenting the time series energy prediction. Lee et al. [80] propose an ultra-short term PV generation forecasting hybrid architecture consisting of a linear sub-model, a Transformer and a LSTM model. The data is decomposed using average pooling, and a linear model is used for the trend component whilst the Transformer and LSTM are used for the residual component of the data. The model is evaluated over short intra-hourly time horizons, and the results show improvement of the PV generation prediction compared to other single (RNN) and hybrid models. PVTransNet is proposed in [99] by Kim et al. for multi-step day-ahead PV generation forecasting. They use different variants of the transformer networks with multiple timeseries input features including historical power generation and weather data. The best performance was achieved by PVTransNet-EDR, a Transformer-based model integrated with a pretrained LSTM that generates weather inputs. The integration of Transformer and LSTM models is highlighted in [82] by Bao et al. to improve the accuracy of the State of Charge estimation in lithium batteries. The transformer model processes the entire set of inputs for long term forecasting. The LSTM model process shorter sequences that are reconstructed based on the Transformer output and the input features. The results show that leveraging both the LSTM and Transformer models leads to improved prediction accuracy. In [90], Wang and Huang propose ConvLTT, a hybrid deep learning model combining ConvLSTM and TCN for short-term PV power forecasting at 5-minute intervals. It learns local and global temporal patterns, enhanced by a novel cross-day data stacking technique that aligns identical time slots across days to improve temporal feature learning. Lara-Benítez et al. [89], short-term solar irradiance forecasting in real-time streaming conditions is explored, by using the ADLStream asynchronous learning framework. Four classic deep learning models are compared using data from a Canadian PV plant. While the Transformer struggled with fast adaptation, MLP and CNN performed better on 30-second forecasts using 3-minute input windows. Song et al. [87] proposes a hybrid RNN–Vanilla Transformer–LSTM model for 1-day ahead solar power forecasting. A concatenated input structure helps with the interpretation of weather variability, while the model shows strong performance across varying weather conditions using data from a 5 MW plant in Pakistan, achieving good results. A transformer-based model for multi-step solar irradiance forecasting up to 10 hours ahead is introduced by Mo et al. [95]. It features a distillation encoder and generative decoder, reducing the computational cost. MSSP demonstrates strong performance using real-world measurements, in diverse weather and real-time conditions, obtaining important cost reductions and profit increases in day-ahead electricity market applications. Bashir et al. [86] propose two hybrid models, the CNN-ABiLSTM and CNN-Transformer-MLP, for photovoltaic and wind power forecasting using quarter-hourly data from Germany and Luxembourg. The CNN-Transformer-MLP model integrates CNN, Transformer encoder with positional encoding, and MLP for short-term prediction. Both proposed models outperform standard baselines, such as CNN or BiLSTM across various metrics. Jiang et al. [96] proposes a dynamic V2X value-stacking framework for EVs in a residential scenario and it integrates V2B, V2G, and energy trading under network constraints and uncertainty. It employs a hybrid Transformer model to forecast building load, PV generation, and EV arrivals. The model outperforms LSTM and quantifies with good accuracy the potential impact of uncorrected forecasting errors. Liao et al. [97] also proposes a dual-gate temporal fusion transformer model for large-scale land surface solar irradiation forecasting. Built on TFT, the approach introduces dual-gate residual and attention blocks. Trained on meteorological and satellite data from Australia, China, and Japan, the model learns well the spatio-temporal patterns across regions, and it outperforms other baseline models. Zhu et al. [94] introduce the U-LSTM-AFT model for hourly solar irradiance forecasting that integrates U-Net-inspired multi-scale feature extraction, LSTM for learning temporal dependencies, and an attention-free transformer (AFT) large scale modeling. The model uses clear-sky index normalization to address irradiance variability, and it achieves better results than benchmark approaches.

**Table 6: Comparison of Transformer-Based Models for Solar Irradiance Forecasting**

| Input Representation | Encoder-Decoder | Attention Mechanism | Architectural Extensions | Work |
|---|---|---|---|---|
| Sky images; weather data | ViT-based encoder; no decoder | Standard scaled dot-product | Image/numeric data gating module | [42] |
| Satellite and ground-based sky images; aggregated irradiance | ViT-based encoder; generative transformer decoder | | | [74] |
| Aggregated PV power data +/- weather data | TCN decoder | | | [90] |
| | Decoder-only | | Teacher forcing multi-step decoding | [89] |
| | RNN and classic encoders; LSTM decoder | | | [87] |
| | Standard encoder-decoder | Multi-timescale fluctuation aggregation (MFA) attention | Contrastive learning module based on similar day selection | [49] |
| Aggregated solar irradiance data +/- weather data | Encoder with distilling; generative decoder | Standard scaled dot-product | Convolutional distillation module | [95] |
| | Encoder-only | | CNN and MLP modules | [86] |
| | GRU encoder–decoder | | | [66] |
| | Informer architecture | ProbSparse attention | Wavelet decomposition module | [39] |

|  | LSTM temporal encoder & static encoder; TFT decoder | Dual-gate multi-head cross attention | Dual gate Gated Residual Network | [97] |
| --- | --- | --- | --- | --- |
|  | LSTM encoder | Predefined weighted aggregation attention |  | [94] |
| **PV Power generation, solar irradiance, temperature, humidity, dust concentration, wind** | Standard Encoder-decoder | Multi-head attention | Linear sub-model for trend component and complementary LSTM sub-model for residual component | [80] |
| **Weather observation** |  |  | Pre-trained LSTM network | [99] |
| **Meteorological (irradiance, temperature, humidity)** |  | Cross-residual attention | Convolutional Weighted Fusion Model | [83] |
| **Meteorological** |  | Self-attention mechanism | Combined proportion-integral-derivative and deep reinforcement learning for smart generation control | [84] |
| **Residential building load, rooftop PV generation, EV arrivals** | GRU and standard encoder; TFT decoder | ProbSparse Self-Attention; Temporal Attention |  | [96] |

## 5. LLMs for energy management

The integration of LLMs into energy systems management represents a significant advancement in addressing the complex challenges of modern energy landscapes [118]. As power grids become increasingly decentralized, intermittent renewable sources proliferate, and demand patterns grow more dynamic, traditional analytical approaches are often insufficient to capture the multi-dimensional complexities involved [119]. LLMs have already been applied in smart energy management, including areas such as energy security and carbon emission prediction. Feng et al. [100] proposed a methodology for predicting daily industrial carbon emissions on a province level by fine-tuning pre-trained language models, specifically the T5 model [120]. Unlike traditional approaches that rely on numerical time-series models, this method leverages the capabilities of natural language processing to generate predictions in textual format. Also, in [101] the potential of LLMs in identifying imminent electrical safety incidents has been demonstrated. Events such as overloads or short circuits are determined by converting numerical electricity usage data into text-based formats suitable for language model input. By training on large-scale datasets from residential households, these models have shown an improved ability to detect early warning signs of hazardous electrical conditions, offering a proactive and scalable approach to risk mitigation. This growing body of work underscores the versatility of LLMs not only in environmental modeling but also in real-time anomaly detection, further establishing their relevance across multiple domains of critical infrastructure monitoring. However, applying LLMs to modern power systems may also incur potential security threats, which have not been fully recognized so far. To this end, Ruan et al. [121] analyzes potential threats incurred by applying LLMs to power systems, emphasizing the need for urgent research and development of countermeasures. Wu et al. [28] have developed a spatio-temporal enhanced pre-trained LLM to overcome the challenges in wind power forecasting. The model applies the reasoning capabilities of LLMs to overcome limitations of traditional forecasting methods by processing decomposed wind series data, incorporating spatial-temporal prompts, and utilizing autoregressive fine-tuning to capture complex patterns in wind speed data across multiple turbines. Similarly, in [29] a spatio-temporal forecasting model based on Bidirectional Encoder Representations from Transformers (BERT) [122] is showcased. This approach leverages BERT's bidirectional attention capabilities to capture intricate spatial and temporal dependencies within wind power data. The model implements specialized encoding designed specifically for the unique characteristics of wind power generation patterns by employing a multi-stage fine-tuning approach that first aligns the language model with spatio-temporal data before optimizing downstream forecasting tasks.

**Table 7: Overview of LLM Applications in the Energy Sector**

| Work | Domain | LLM | Fine tuning method | Tasks per LLM agent |
|---|---|---|---|---|
| **Non-Agentic** | | | | |
| [100] | Industrial carbon emissions | T5 | Partial autocorrelation | |
| [101] | Electrical safety incidents in residential meters | Custom-made | Chain-of-thought | |
| [28] | Wind speed forecasting | GPT, BERT, T5, GPT-2 | Positional embeddings | |
| [29] | | BERT | Low-rank adaptation | |
| [105] | Load Forecasting | TimeGPT | Zero shot learning; Few-shot learning | |
| [102] | BEM | GPT-4o | Feature engineering | |
| [123] | | T5 | Byte-pair encoding | |
| [32] | Building load forecasting | GPT-3.5 | Optimal prompts | |
| [113] | Building management | GPT | Occupant interaction | |
| [31] | Intrusion Detection for Communication protocols | ChatGPT 4.0, Claude 2, Bard, PaLM | Prompt Engineering and human in the loop training | |
| [115] | Testing Intrusion Detection System against cyberattacks | Conditional Tabular Generative Adversarial Network | Trained using a generated adversarial attack dataset | |
| [116] | Sky images prediction for PV generation forecasting | SkyGPT (Physics-constrained VideoGPT) | Training on image datasets and incorporated physics informed | |
| [117] | Wind power forecasting for net-zero carbon emissions | XAI-driven generative model | Probabilistic Inference, t-SNE and DBSCAN for interpretability | |
| **Agentic** | | | | |
| [33] | BIM/Retrofit recommendation | GPT 4-Turbo | Supervised fine-tuning / Prompt engineering | Information extraction & processing, Performance diagnosis, Retrofit recommendation |

| [108] |  | GPT | EnergyPlus engineering reference | Information extraction, IDF Generation, Debugging |
|---|---|---|---|---|
| [106] | BEM | GPT-4, Gemini | Single IDF object | Pre-processing & information extraction, IDF Generation, Debugging |
| [109] | EV manufacturing | GPT-4o | Prompt engineering | Information extraction & processing, Decision support |
| [30] | Energy Grid Management | BERT | Token classification using LEAP-based dataset | Provides information needed by human operators |

Researchers in the domain of sustainable urban development have also begun to explore the integration of LLMs into complex modeling tasks, such as BEM. A recent study introduces GPT-Urban BEM, a GPT-4o-powered framework designed to tackle traditional challenges in urban BEM by leveraging natural language understanding for advanced data interpretation and optimization [102]. Across four case studies involving over 2,000 buildings in Seoul and 31 in Gangwon-do, the model demonstrated proficiency in tasks such as urban data analytics, energy prediction, feature engineering, and energy signature analysis. The results reveal GPT-UBEM's strength in synthesizing heterogeneous datasets and improving the accuracy and interpretability of energy models, all while offering actionable insights for urban planners and policymakers. In this context, multiple studies have introduced fusion techniques of LLMs and EnergyPlus [108], [123], one of the most well-known BEM simulation tools. In [108] the emphasizes is on how LLMs can assist in a range of simulation-related tasks such as input generation and output analysis to error detection, co-simulation, and optimization. Through three case studies, the authors illustrate the model's capacity to automate complex modeling workflows, reduce manual engineering effort, and improve simulation outcomes. The research highlights that effective deployment of LLMs in such technical domains depends heavily on selecting the appropriate modeling techniques and aligning them with engineering needs. Jiang et al. [123], introduces Eplus-LLM, a fine-tuned T5-based platform designed to automate the translation of natural language building descriptions into EnergyPlus models. By tailoring the LLM to interpret user prompts, regardless of tone, misspellings, or incomplete information, the platform enables users without deep expertise in building science or simulation tools to generate accurate, simulation-ready BEM files. The system not only automates model generation but also leverages the EnergyPlus API to run simulations and return performance metrics.

Following the trend to explore agentic workflows to fully automate procedures [124], Zhang et al. [106] present a planning-based LLM workflow that transforms natural language building descriptions into accurate, error-free EnergyPlus models. This framework is structured around four core agents: preprocessing, object extraction, object generation, and debugging. Each agent is designed to break down the complex BEM process into modular, manageable subtasks. Through a case study on the iUnit modular building at the National Renewable Energy Laboratory, the approach outperformed manual methods, naive prompting, and alternative LLM-based workflows in terms of accuracy, reliability, and time efficiency. Furthermore, in [33] a pipeline is proposed consisting of building information processing, performance diagnosis, and retrofit recommendation. Domain-informed LLM agents fulfill the roles of planner, researcher, and advisor. By incorporating external knowledge databases and using retrieval tools for contextual grounding, the framework effectively extracts actionable insights from diverse, unstructured inputs. Those approaches underscore the growing role of LLM-based multi-agent systems as generalizable, trustworthy task solvers capable of significantly reducing human labor in the building sector, while accelerating sustainable energy practices. Liu et al. [109] propose a blockchain-based, LLM-driven scheduling framework tailored to the multi-agent manufacturing of new energy vehicles, aligning with circular economy principles. The manufacturing resources across geographically dispersed factory nodes are virtualized into intelligent agents, and energy-efficient scheduling is optimized using an LLM trained on diverse production data. Unlike traditional methods prone to local optima and long training cycles, the LLM leverages its pre-trained knowledge for self-adaptive decision-making that reduces energy usage and production delays. To address the issue of untrusted or inconsistent production data, a credit evaluation-based consensus mechanism is introduced via blockchain, enhancing data reliability, transparency, and traceability. Zhang et al. [32] proposes a GPT-based automated data-driven building energy load forecasting method that streamlines the process of generating predictive code by addressing key usability and accuracy challenges. Recognizing the difficulty inexperienced users face in crafting effective prompts, the authors introduce automated prompting functions, enhanced through Bayesian optimization, to fine-tune prompt generation for improved prediction performance. Additionally, they incorporate external knowledge bases to enhance code correctness and a self-correction strategy that allows GPT-3.5 to iteratively fix its own errors. Following the development of automated prompt engineering for building energy load forecasting, [113] expands the application of GPT-based systems by integrating them into holistic smart building management platforms. A real-time, IoT-driven framework that simultaneously monitors and manages health, energy consumption, and thermal comfort, incorporating a GPT-powered conversational suggestion system is introduced. By combining IoT-based BIM, cloud simulation, and real-time data streams, the platform facilitates continuous assessment and dynamic adjustment of indoor conditions.

LLMs can be used to improve grid management and optimization either by generating relevant information needed in the management and optimization process, or by generating synthetic and various scenarios to better understand and evaluate the technologies involved. Gamage et al. [30] address the complexities of autonomous energy management considering the dynamic data spaces involved in the decision-making process. They used rule-based language processing together with LLM to enable a conversational agent (chatbot) that also considers providing the information needed by human operators in the decision-making process. Thus, the process of inference over dynamic and various data spaces, and selecting relevant information is simplified. Generative AI has been used in smart grids management for generating and simulating different testing scenarios, involving costly technologies or situations with insufficient data. Asimopoulos et al. [115] use Fast Gradient Sign Method to generate an initial adversarial dataset. This data is used to train a Conditional Tabular Generative Adversarial Network to generate adversarial attacks against AI models that detect intrusion. Using the generated attacks scenarios, the resilience of different AI models against intrusion attacks can be evaluated more extensively. Also, it can be used to assess the impact of different technologies on the smart grid. A generative model is used in [117] to simulate different scenarios for wind power forecasts to show how offshore wind power can help the petrochemical industry achieve net-zero carbon emissions. LLMs can be leveraged as an intrusion detection system for enhancing the cybersecurity of smart grid applications. Zaboli et al. [31] propose to fine-tune LLMs through human interaction and informed by cybersecurity guidelines to detect anomalies in communication protocols. Intrusion detection algorithms are converted to text, enabling more effective LLM training. Multiple LLMs were compared for detecting the anomalies, and ChatGPT 4.0 outperformed the others. Nie et al. [116] propose SkyGPT, an LLM that generates future images of clouds to improve the PV generation prediction. The generation model is physics-constrained, and the evaluation results show that it can generate accurate future images of the sky. Also, using the generated images as input for a U-Net prediction model the power generation predictions accuracy was significantly improved. Finally, in [105] an LLM-based solution for load time series prediction where the generative transformer is pre-trained on large time series datasets from various fields. Then, the TimeGPT model can be fine-tuned using load dataset, especially in cases where the available historical load data is limited. The evaluation results show that the TimeGPT can outperform benchmark neural network-based prediction models over most datasets for short time predictions. However, a specific dataset can have particularities, different distributions of the time series and in these cases the performance can be affected.

## 6. Perspectives and challenges

The review presented in this paper highlights the transformative potential of Transformer-based models and the shift towards LLMs across diverse energy system applications, from forecasting and optimization to anomaly detection and decision support. As GenAI models continue to evolve, their capabilities in handling temporal dependencies, multimodal inputs, and complex reasoning tasks are progressively harnessed within the energy sector. However, realizing their full potential in the context of smart grids and DT systems requires not only technical advancements but also a rethink of how intelligence is embedded in energy management frameworks. In this section we discuss opportunities and open challenges associated with the integration of LLMs into energy-aware, adaptive, and agentic DTs.

Moving forward, transformers and LLM-based solutions will continue to evolve, with increased attention on scalability, adaptability, and energy grid management specific adaptations. Emerging trends include zero-shot and few-shot learning capabilities, allowing LLMs to be integrated and perform effectively with energy management scenarios with minimal fine-tuning [125]. Moreover, advancements in computational efficiency through model compression, and energy-aware model designs will likely streamline smart grid deployments enabling their orchestration across the entire edge-fog-cloud computational continuum [126]. LLM's pruning and distillation will make it possible to deploy these models on edge devices or substations closer to the energy prosumption site.

Fine-tuned LLMs are already starting to be integrated in energy grid management tasks, however significant challenges are clustered around making interactive, adaptive, and situationally aware of real-time stimuli [127]. LLM agents have already been coined in the state-of-the-art, as autonomous, goal-driven modules that are capable of monitoring, adapting and coordinating the tasks execution being relevant for smart grid management decision making [128]. At the same time DTs are seen as the technological enablers of the next-generation smart grid management solutions [129]. Their hybrid approach of integrating physical realities and components model simulations manage to capture finer-grained details and provide more precise predictions and enhanced decision making for efficient development, integration and operation of expensive energy assets. In this context, we consider that the integration of LLMs and GenAI has the potential of transforming DT into agentic models with autonomy, proactivity, and conversational capabilities able to interact and cooperate with other DTs in the management of the energy grid (see Figure 8).

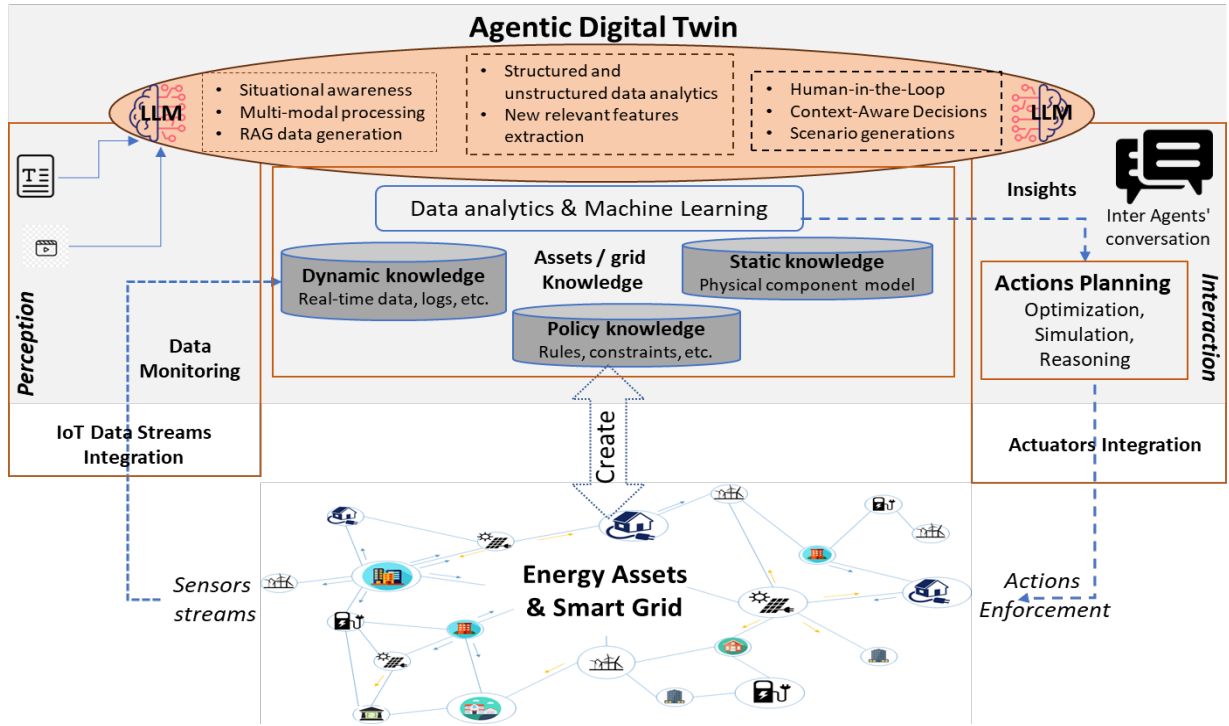

**Figure 8.** Architecture of a smart grid Agentic DT showing the where the LLMs can augment and support the management processes

In our vision, the DTs will depart from nowadays monitoring processes that aim of observing energy system states or environments through sensors or data sources towards interpreting monitored data to understand the context or situation. As result a meaningful model of energy assets or system will be created, enabling energy awareness and situational understanding. This process is accelerated by LLMs integration to enhance monitored data towards multimodal sources and fusing it into dynamic situational knowledge while integrate in a various data retrieval mechanism [108]. LLM may transform raw sensory data into meaningful knowledge representations of the energy assets and systems supporting future energy management scenario simulation. They can process and understand textual description of the energy asset operation and they may be used to better understand the energy time series data [106]. For example, LLMs can parse manuals, specifications, and reports to extract key attributes like capacity, operating thresholds and generate structured metadata on assets capabilities. Moreover, this may offer possible operational explanations for some time series energy data anomalies. However, LLMs need to be adapted or fine-tuned to understand the energy domain specific context which is a time-consuming and computationally expensive.

Due to LLM integration DTs will access and integrate information from vast external sources, enabling autonomous information retrieval either through RAG or prompts facilitating better perception of energy asset changes or real-time context-awareness [130]. However, improved attention mechanisms need to be defined to ensure relevance and focus on the data used. For example, LLMs equipped with APIs calling functions may enable DTs to identify external tools or new databases (e.g. energy policies, statistics, etc.) that need to be queried while maintaining the awareness of the energy management process. Additionally, LLMs can support the generation of missing data by retrieving relevant information using semantic search via embeddings that offer contextualized multimodal feeds leading to better-grounded energy management decision making. For example, leveraging appliance manuals for online and residents' habits extracted from smart home log files the LLMs can describe data patterns that are missing or may generate synthetic time series data.

The perception challenges are related to the updating or re-training the LLM with the new data streams, which is a rather difficult and computational costly process [128]. Therefore, in time LLMs may provide outdated or obsolete responses limiting their usefulness for smart grid energy management applications that are rather in time sensitive.

In terms of data analysis, the generalization abilities of LLMs in few-shot learning may significantly improve the adaptability of agentic DT enabling them to analyze diverse and heterogenous data without extensive task-specific training [131]. This will enable the construction of more generic smart grid DT agents automatizing the adaptation to different energy management tasks. This level of adaptability could be highly beneficial in data-scarce energy management scenarios with multiple coordinated distributed energy resources. LLMs offer native support for multimodal inputs and multitask learning, enabling the combination between diverse datasets such as numerical energy readings, weather forecasts specified in text, satellite images, or news and market reports [132]. Therefore, it will enhance the data analytics of smart grid DTs, improving contextual understanding and uncovering hidden insights.

The deployment of AI-based forecasting models has the potential of significantly improving decision-making for market participants and operators, playing an essential role in the current operational scenario characterized by high shares of renewable energy generation. AI-based forecasting models may increase the effectiveness of energy bidding strategies under uncertainty, reducing the microgrid operational risks. LLMs will further improve this by extracting relevant features from textual logs or metadata that can be incorporated into machine learning algorithms, and predictive analytics, to facilitate better and more accurate forecasting processes [133]. Moreover, LLM can bridge the gap between structured time series data and unstructured text information. For example, in case of a voltage dip, an LLM can correlate it with a weather report saying strong winds and thunderstorms in the area by reading event logs and matching them to patterns in structured data. They may augment and improve the non-intrusive load monitoring processes [33] behind the meter energy assets by enabling the correlation between the sensor reading (i.e. power spikes at specific hours) with device logs saying that certain appliances had been stared, while leveraging on the RAG to read appliances manuals for accurate missing data generation.

The analytics' related challenges are driven by data quality and biases in data that may lead to analytics biases. The models are trained on energy domain datasets that may be non-representative, with anomalies or incorrectly labelled generating additional performance challenges on new, unseen data or unseen energy assets of system configurations, where unpredictable conditions may arise.

The DTs' decision-making process is driven by the interaction with the real-world environment either directly or simulated. It usually involves building an action space, selecting relevant actions and constructing an action plan which is then optimized based on feedback or learning policies. LLMs have good capabilities in interpreting objectives and generating plausible actions, therefore, they can contribute in each of these phases supporting the implementation of agentic DTs [109]. LLMs can generate a wide variety of possible energy management actions from natural language descriptions related to energy systems considering domain specific knowledge. For example, they support DTs in planning the ramping down of non-essential loads for off-peak hours while suggesting demand-response signals to reduce peak electricity load in a region due to the low availability of renewables. LLM can provide detailed and coherent energy management action plans by interpreting the objectives or desired outcomes while analyzing context and extracting key information [127]. For example, they can support DTs in planning of fast-ramping energy units as spinning reserve and optimizing energy storage system to dispatch profiles while considering the weather forecast new in specific region. LLM can help in dividing complex tasks into manageable steps, organizing them sequentially or hierarchically based on logical dependencies.

Using LLM for constructing a meaningful and bounded set of actionable energy management operations that reflect smart grid constraints is still challenging. They are exposed to hallucinations therefore may generate plausible but irrelevant or infeasible management actions. As potential mitigation LLM needs to be combined with energy asset knowledge and energy domain simulation engines [134]. Additionally, they need to adapt and improve based on real world feedback, performance data, or learning policies which is a complex and computational costly process. Finally, LLMs can help validate whether a DTs action is safe, or feasible by simulating its impact or checking against smart grid or energy resources constraints [135]. However, LLMs cannot execute or enforce actions directly in the smart grid therefore they cannot be used for direct real-time control of energy systems. Their role should remain to advise and support decision making, relying on human or connected devices to carry out action enforcement. Therefore, humans will remain a key in the decision-making loop for energy management.

## 7. Conclusions

This review analysed 99 publications from 2021 to 2025 on the use of Transformer models and LLMs in the energy sector. Transformer-based architectures—especially Temporal Fusion Transformer, Informer, and their variants have proven to be effective in short and long-term forecasting of electricity demand, thermal loads, and renewable generation, consistently reducing error metrics compared to LSTM, GRU, and CNN baselines. Important architectural improvements identified include spatial-temporal attention, patch-based decomposition, and hybrid physics-informed models. LLMs, though newer in the domain, are rapidly extending their impact. Fine-tuned models like GPT-3.5, GPT-4o, and domain-specific variants (e.g., TimeGPT, BERT4ST) have been applied in building energy modelling, scenario generation, anomaly detection, and decision support. Their strength lies in integrating structured time series data with unstructured text (e.g., manuals, logs, weather reports) and enabling reasoning across modalities. Thus, the main findings are summarized as the following: (a) Transformer models now dominate energy forecasting tasks due to their adaptability and accuracy; (b) LLMs extend capabilities beyond prediction—enabling integration of knowledge, user interaction, and scenario synthesis; and (c) Early examples show potential for multi-agent LLM workflows automating complex energy modelling tasks.

In terms of research perspectives, the central insight of this paper is the emerging role of LLMs in transforming DTs into Agentic Digital Twins: autonomous, proactive, and communicative models capable of not only mirroring physical assets but featuring situational awareness and reasoning capabilities for planning and recommending energy management actions. The key challenges ahead were summarised to the following points: (a) the high cost of domain adaptation and continuous fine-tuning; (b) the risk of hallucinations and lack of action validation in operational contexts; (c) the need for real-time, on-device inference in edge or substation environments; and finally (d) the current lack of standards for integrating LLMs in safety-critical decision loops.

The path forward involves building domain-adapted, computationally efficient, and interpretable LLMs that complement robust forecasting models and support actionable, human-supervised decision-making in smart grids. Agentic DTs will not replace operators but enhance their capability to manage increasingly complex energy systems.

**Acknowledgement**

This work was supported by the project "Romanian Hub for Artificial Intelligence-HRIA", Smart Growth, Digitization and Financial Instruments Program, MySMIS no. 334906. The research has been partially supported by the European Union's Horizon Europe research and innovation programme under Grant Agreements number 101136216 (Hedge-IoT) and 101103998 (DEDALUS). Views and opinions expressed are, however, those of the author(s) only and do not necessarily reflect those of the European Union or the European Climate, Infrastructure, and Environment Executive Agency. Neither the European Union nor the granting authority can be held responsible for them.

**References**

[1] Baidya, S., Potdar, V., Ray, P. P., & Nandi, C. (2021). Reviewing the opportunities, challenges, and future directions for the digitalization of energy. Energy Research & Social Science, 81, 102243.
[2] Xie, L., Zheng, X., Sun, Y., Huang, T., & Bruton, T. (2022). Massively digitized power grid: Opportunities and challenges of use-inspired AI. Proceedings of the IEEE, 111(7), 762-787.
[3] Monaco, R., Bergaentzlé, C., Vilaplana, J. A. L., Ackom, E., & Nielsen, P. S. (2024). Digitalization of power distribution grids: Barrier analysis, ranking and policy recommendations. Energy Policy, 188, 114083.
[4] Filonchyk, M., Peterson, M. P., Yan, H., Gusev, A., Zhang, L., He, Y., & Yang, S. (2024). Greenhouse gas emissions and reduction strategies for the world's largest greenhouse gas emitters. Science of The Total Environment, 944, 173895.
[5] Bataille, C., Åhman, M., Neuhoff, K., Nilsson, L. J., Fischedick, M., Lechtenböhmer, S., ... & Rahbar, S. (2018). A review of technology and policy deep decarbonization pathway options for making energy-intensive industry production consistent with the Paris Agreement. Journal of Cleaner Production, 187, 960-973.
[6] Li, J., Herdem, M. S., Nathwani, J., & Wen, J. Z. (2023). Methods and applications for Artificial Intelligence, Big Data, Internet of Things, and Blockchain in smart energy management. Energy and AI, 11, 100208.
[7] Sarmas, E. Artificial Intelligence for Energy Systems: Driving Intelligent, Flexible and Optimal Energy Management. Springer Nature.
[8] Entezari, A., Aslani, A., Zahedi, R., & Noorollahi, Y. (2023). Artificial intelligence and machine learning in energy systems: A bibliographic perspective. Energy Strategy Reviews, 45, 101017.
[9] Liu, Z., Sun, Y., Xing, C., Liu, J., He, Y., Zhou, Y., & Zhang, G. (2022). Artificial intelligence powered large-scale renewable integrations in multi-energy systems for carbon neutrality transition: Challenges and future perspectives. Energy and AI, 10, 100195.
[10] Han, K., Xiao, A., Wu, E., Guo, J., Xu, C., & Wang, Y. (2021). Transformer in transformer. Advances in neural information processing systems, 34, 15908-15919.
[11] Nascimento, E. G. S., de Melo, T. A., & Moreira, D. M. (2023). A transformer-based deep neural network with wavelet transform for forecasting wind speed and wind energy. Energy, 278, 127678.
[12] Xu, C., & Chen, G. (2024). Interpretable transformer-based model for probabilistic short-term forecasting of residential net load. International Journal of Electrical Power & Energy Systems, 155, 109515.
[13] Liu, J., & Fu, Y. (2023). Renewable energy forecasting: A self-supervised learning-based transformer variant. Energy, 284, 128730.
[14] Antonesi, G., Cioara, T., Anghel, I., Papias, I., Michalakopoulos, V., & Sarmas, E. (2025). Hybrid transformer model with liquid neural networks and learnable encodings for buildings' energy forecasting. Energy and AI, 20, 100489.
[15] Achiam, J., Adler, S., Agarwal, S., Ahmad, L., Akkaya, I., Aleman, F. L., ... & McGrew, B. (2023). Gpt-4 technical report. arXiv preprint arXiv:2303.08774.
[16] Chowdhery, A., Narang, S., Devlin, J., Bosma, M., Mishra, G., Roberts, A., ... & Fiedel, N. (2023). Palm: Scaling language modeling with pathways. Journal of Machine Learning Research, 24(240), 1-113.
[17] Garza, A., Challu, C., & Mergenthaler-Canseco, M. (2023). TimeGPT-1. arXiv preprint arXiv:2310.03589.

[18] Yi, H., Zhang, S., An, D., & Liu, Z. (2024). PatchesNet: PatchTST-based multi-scale network security situation prediction. Knowledge-Based Systems, 299, 112037.
[19] Siyuan Huang, Yepeng Liu, Haoyi Cui, Fan Zhang, Jinjiang Li, Xiaofeng Zhang, Mingli Zhang, Caiming Zhang, MEAformer: An all-MLP transformer with temporal external attention for long-term time series forecasting, Information Sciences, Volume 669, 2024, 120605, ISSN 0020-0255, https://doi.org/10.1016/j.ins.2024.120605.
[20] Zheng Wang, Haowei Ran, Jinchang Ren, Meijun Sun, PWDformer: Deformable transformer for long-term series forecasting, Pattern Recognition, Volume 147, 2024, 110118, ISSN 0031-3203, https://doi.org/10.1016/j.patcog.2023.110118.
[21] Jun Wei Chan, Chai Kiat Yeo, A Transformer based approach to electricity load forecasting, The Electricity Journal, Volume 37, Issue 2, 2024, 107370, ISSN 1040-6190, https://doi.org/10.1016/j.tej.2024.107370.
[22] Pengfei Zhao, Weihao Hu, Di Cao, Zhenyuan Zhang, Wenlong Liao, Zhe Chen, Qi Huang, Enhancing multivariate, multi-step residential load forecasting with spatiotemporal graph attention-enabled transformer, International Journal of Electrical Power & Energy Systems, Volume 160, 2024, 110074, ISSN 0142-0615, https://doi.org/10.1016/j.ijepes.2024.110074.
[23] Faisal Saeed, Abdul Rehman, Hasnain Ali Shah, Muhammad Diyan, Jie Chen, Jae-Mo Kang, SmartFormer: Graph-based transformer model for energy load forecasting, Sustainable Energy Technologies and Assessments, Volume 73, 2025, 104133, ISSN 2213-1388, https://doi.org/10.1016/j.seta.2024.104133.
[24] Miguel López Santos, Saúl Díaz García, Xela García-Santiago, Ana Ogando-Martínez, Fernando Echevarría Camarero, Gonzalo Blázquez Gil, Pablo Carrasco Ortega, Deep learning and transfer learning techniques applied to short-term load forecasting of data-poor buildings in local energy communities, Energy and Buildings, Volume 292, 2023, 113164, ISSN 0378-7788, https://doi.org/10.1016/j.enbuild.2023.113164.
[25] Wei Zhang, Hongyi Zhan, Hang Sun, Mao Yang, Probabilistic load forecasting for integrated energy systems based on quantile regression patch time series Transformer, Energy Reports, Volume 13, 2025, Pages 303-317, ISSN 2352-4847, https://doi.org/10.1016/j.egyr.2024.11.057.
[26] Jingfei Wang, Danya Xu, Yuanzheng Li, Mohammad Shahidehpour, Tao Yang, Transformer-based probabilistic demand forecasting with adaptive online learning, Electric Power Systems Research, Volume 240, 2025, 111255, ISSN 0378-7796, https://doi.org/10.1016/j.epsr.2024.111255.
[27] Junlong Tong, Liping Xie, Wankou Yang, Kanjian Zhang, Junsheng Zhao, Enhancing time series forecasting: A hierarchical transformer with probabilistic decomposition representation, Information Sciences, Volume 647, 2023, 119410, ISSN 0020-0255, https://doi.org/10.1016/j.ins.2023.119410.
[28] Tangjie Wu, Qiang Ling, STELLM: Spatio-temporal enhanced pre-trained large language model for wind speed forecasting, Applied Energy, Volume 375, 2024, 124034, ISSN 0306-2619, https://doi.org/10.1016/j.apenergy.2024.124034.
[29] Zefeng Lai, Tangjie Wu, Xihong Fei, Qiang Ling, BERT4ST:: Fine-tuning pre-trained large language model for wind power forecasting, Energy Conversion and Management, Volume 307, 2024, 118331, ISSN 0196-8904, https://doi.org/10.1016/j.enconman.2024.118331.
[30] G. Gamage, N. Mills, P. Rathnayaka, A. Jennings and D. Alahakoon, "Cooee: An Artificial Intelligence Chatbot for Complex Energy Environments," 2022 15th International Conference on Human System Interaction (HSI), Melbourne, Australia, 2022, pp. 1-5, doi: 10.1109/HSI55341.2022.9869464.
[31] A. Zaboli, S. L. Choi, T. -J. Song and J. Hong, "ChatGPT and Other Large Language Models for Cybersecurity of Smart Grid Applications," 2024 IEEE Power & Energy Society General Meeting (PESGM), Seattle, WA, USA, 2024, pp. 1-5, doi: 10.1109/PESGM51994.2024.10688863..
[32] Chaobo Zhang, Jian Zhang, Yang Zhao, Jie Lu, Automated data-driven building energy load prediction method based on generative pre-trained transformers (GPT), Energy, Volume 318, 2025, 134824, ISSN 0360-5442, https://doi.org/10.1016/j.energy.2025.134824.
[33] Tong Xiao, Peng Xu, Exploring automated energy optimization with unstructured building data: A multi-agent based framework leveraging large language models, Energy and Buildings, Volume 322, 2024, 114691, ISSN 0378-7788, https://doi.org/10.1016/j.enbuild.2024.114691.
[34] Page, M.J., McKenzie, J.E., Bossuyt, P.M. et al. The PRISMA 2020 statement: an updated guideline for reporting systematic reviews. Syst Rev 10, 89 (2021). https://doi.org/10.1186/s13643-021-01626-4
[35] Page M J, Moher D, Bossuyt P M, Boutron I, Hoffmann T C, Mulrow C D et al. PRISMA 2020 explanation and elaboration: updated guidance and exemplars for reporting systematic reviews BMJ 2021; 372 :n160 doi:10.1136/bmj.n160

[36] Li, K., Rollins, J. & Yan, E. Web of Science use in published research and review papers 1997–2017: a selective, dynamic, cross-domain, content-based analysis. Scientometrics 115, 1–20 (2018). https://doi.org/10.1007/s11192-017-2622-5
[37] H. Zhao, Y. Wu, L. Ma and S. Pan, "Spatial and Temporal Attention-Enabled Transformer Network for Multivariate Short-Term Residential Load Forecasting," in IEEE Transactions on Instrumentation and Measurement, vol. 72, pp. 1-11, 2023, Art no. 2524611, doi: 10.1109/TIM.2023.3305655
[38] Wang, Yun; Xu, Houhua; Song, Mengmeng; Zhang, Fan; Li, Yifen; Zhou, Shengchao; Zhang, Lingjun, A convolutional Transformer-based truncated Gaussian density network with data denoising for wind speed forecasting, APPLIED ENERGY, Volume 333, 1 March 2023, 120601
[39] Zhang, Hanjin; Li, Bin; Su, Shun-Feng; Yang, Wankou; Xie, Liping, A Novel Hybrid Transformer-Based Framework for Solar Irradiance Forecasting Under Incomplete Data Scenarios, IEEE TRANSACTIONS ON INDUSTRIAL INFORMATICS, 2024
[40] Li, Long; Su, Xingyu; Bi, Xianting; Lu, Yueliang; Sun, Xuetao, A novel Transformer-based network forecasting method for building cooling loads, ENERGY AND BUILDINGS, 2023
[41] Dong, Changyin; Xiong, Zhuozhi; Zhang, Chu; Li, Ni; Li, Ye; Xie, Ning; Zhang, Jiarui; Wang, Hao, A transformer-based approach for deep feature extraction and energy consumption prediction of electric buses based on driving distances, APPLIED ENERGY, 2025
[42] Zhang, Liwenbo; Wilson, Robin; Sumner, Mark; Wu, Yupeng, Advanced multimodal fusion method for very short-term solar irradiance forecasting using sky images and meteorological data: A gate and transformer mechanism approach, RENEWABLE ENERGY, 2023
[43] Huang, Siyuan; Liu, Yepeng; Zhang, Fan; Li, Yue; Li, Jinjiang; Zhang, Caiming, CrossWaveNet: A dual-channel network with deep cross-decomposition for Long-term Time Series Forecasting, EXPERT SYSTEMS WITH APPLICATIONS, 2024
[44] Zhou, Chengjie; Che, Chao; Wang, Pengfei; Zhang, Qiang, Diformer: A dynamic self-differential transformer for new energy power autoregressive prediction, KNOWLEDGE-BASED SYSTEMS, 2023
[45] Rong, Jing; Wang, Cong; Zhou, Qiuzhan; He, Yunxue; Wu, Huinan, Enhancing non-intrusive load monitoring through transfer learning with transformer models, ENERGY AND BUILDINGS, 2025
[46] Huang, Siyuan; Liu, Yepeng, FL-Net: A multi-scale cross-decomposition network with frequency external attention for long-term time series forecasting, KNOWLEDGE-BASED SYSTEMS, 2024
[47] Wang, Chao-fan; Liu, Kui-xing; Peng, Jieyang; Li, Xiang; Li, Xiu-feng; Zhang, Jia-wan; Niu, Zhi-bin, High-precision energy consumption forecasting for large office building using a signal decomposition-based deep learning approach, ENERGY, 2025
[48] Su, Jing; Xie, Dirui; Duan, Yuanzhi; Zhou, Yue; Hu, Xiaofang; Duan, Shukai, MDCNet: Long-term time series forecasting with mode decomposition and 2D convolution, KNOWLEDGE-BASED SYSTEMS, 2024
[49] Yuan, Liang; Wang, Xiangting; Sun, Yao; Liu, Xubin; Dong, Zhao Yang, Multistep photovoltaic power forecasting based on multi-timescale fluctuation aggregation attention mechanism and contrastive learning, INTERNATIONAL JOURNAL OF ELECTRICAL POWER & ENERGY SYSTEMS, 2025
[50] Han, Yongming; Han, Longkun; Shi, Xinwei; Li, Jun; Huang, Xiaoyi; Hu, Xuan; Chu, Chong; Geng, Zhiqiang, Novel CNN-based transformer integrating Boruta algorithm for production prediction modeling and energy saving of industrial processes, EXPERT SYSTEMS WITH APPLICATIONS, 2024
[51] Mo, Site; Wang, Haoxin; Li, Bixiong; Xue, Zhe; Fan, Songhai; Liu, Xianggen, Powerformer: A temporal-based transformer model for wind power forecasting, ENERGY REPORTS, 2024
[52] Wu, Qiang; Zheng, Hongling; Guo, Xiaozhu; Liu, Guangqiang, Promoting wind energy for sustainable development by precise wind speed prediction based on graph neural networks, RENEWABLE ENERGY, 2022
[53] Yu, Yang; Ma, Ruizhe; Ma, Zongmin, Robformer: A robust decomposition transformer for long-term time series forecasting, PATTERN RECOGNITION, 2024
[54] Zhang, Qingyong; Zhou, Shiyang; Xu, Bingrong; Li, Xinran, TCAMS-Trans: Efficient temporal-channel attention multi-scale transformer for net load forecasting, COMPUTERS & ELECTRICAL ENGINEERING, 2024
[55] Yu, Die; Liu, Tong; Wang, Kai; Li, Kang; Mercangöz, Mehmet; Zhao, Jian; Lei, Yu; Zhao, Ruofan, Transformer based day-ahead cooling load forecasting of hub airport air-conditioning systems with thermal energy storage, ENERGY AND BUILDINGS, 2024

[56] Yuan, Jun; Chen, Si-Zhe; Yu, Samson S.; Zhang, Guidong; Chen, Zhe; Zhang, Yun, A Kernel-Based Real-Time Adaptive Dynamic Programming Method for Economic Household Energy Systems, IEEE TRANSACTIONS ON INDUSTRIAL INFORMATICS, 2023

[57] Wang, Chen; Wang, Ying; Ding, Zhetong; Zheng, Tao; Hu, Jiangyi; Zhang, Kaifeng, A Transformer-Based Method of Multienergy Load Forecasting in Integrated Energy System, IEEE TRANSACTIONS ON SMART GRID, 2022

[58] Gao, Jiaxin; Chen, Yuntian; Hu, Wenbo; Zhang, Dongxiao, An adaptive deep-learning load forecasting framework by integrating transformer and domain knowledge, ADVANCES IN APPLIED ENERGY, 2023

[59] Yan, Qin; Lu, Zhiying; Liu, Hong; He, Xingtang; Zhang, Xihai; Guo, Jianlin, An improved feature-time Transformer encoder-Bi-LSTM for short-term forecasting of user-level integrated energy loads, ENERGY AND BUILDINGS, 2023

[60] Ozen, Serkan; Yazici, Adnan; Atalay, Volkan, Hybrid deep learning models with data fusion approach for electricity load forecasting, EXPERT SYSTEMS, 2025

[61] Cui, Yunlong; Li, Zhao; Wang, Yusong; Dong, Danhuang; Gu, Chenlin; Lou, Xiaowei; Zhang, Peng, Informer model with season-aware block for efficient long-term power time series forecasting, COMPUTERS & ELECTRICAL ENGINEERING, 2024

[62] Cen, Senfeng; Lim, Chang Gyoon, Multi-Task Learning of the PatchTCN-TST Model for Short-Term Multi-Load Energy Forecasting Considering Indoor Environments in a Smart Building, IEEE ACCESS, 2024

[63] Andréia B.A. Ferreira, Jonatas B. Leite, Denis H.P. Salvadeo, Power substation load forecasting using interpretable transformer-based temporal fusion neural networks, Electric Power Systems Research, Volume 238, 2025, 111169, ISSN 0378-7796, https://doi.org/10.1016/j.epsr.2024.111169.

[64] Pham Canh Huy; Nguyen Quoc Minh; Nguyen Dang Tien; Tao Thi Quynh Anh, Short-Term Electricity Load Forecasting Based on Temporal Fusion Transformer Model, IEEE ACCESS, 2022

[65] Giacomazzi, Elena; Haag, Felix; Hopf, Konstantin, Short-Term Electricity Load Forecasting Using the Temporal Fusion Transformer: Effect of Grid Hierarchies and Data Sources, PROCEEDINGS OF THE 2023 THE 14TH ACM INTERNATIONAL CONFERENCE ON FUTURE ENERGY SYSTEMS, E-ENERGY 2023, 2023

[66] Alorf, Abdulaziz; Khan, Muhammad Usman Ghani, Solar Irradiance Forecasting Using Temporal Fusion Transformers, INTERNATIONAL JOURNAL OF ENERGY RESEARCH, 2025

[67] Du, Pei; Ye, Yuxin; Wu, Han; Wang, Jianzhou, Study on deterministic and interval forecasting of electricity load based on multi-objective whale optimization algorithm and transformer model, EXPERT SYSTEMS WITH APPLICATIONS, 2025

[68] Semmelmann, Leo; Hertel, Matthias; Kircher, Kevin J.; Mikut, Ralf; Hagenmeyer, Veit; Weinhardt, Christof, The impact of heat pumps on day-ahead energy community load forecasting, APPLIED ENERGY, 2024

[69] Zhang, Qingyong; Chen, Jiahua; Xiao, Gang; He, Shangyang; Deng, Kunxiang, TransformGraph: A novel short-term electricity net load forecasting model, ENERGY REPORTS, 2023

[70] Lin, Zijie; Xie, Linbo; Zhang, Siyuan, A compound framework for short-term gas load forecasting combining time-enhanced perception transformer and two-stage feature extraction, ENERGY, 2024

[71] Iqbal, Muhammad Sajid; Adnan, Muhammad, Edge computing and transfer learning-based short-term load forecasting for residential and commercial buildings, ENERGY AND BUILDINGS, 2025

[72] Zheng, Peijun; Zhou, Heng; Liu, Jiang; Nakanishi, Yosuke, Interpretable building energy consumption forecasting using spectral clustering algorithm and temporal fusion transformers architecture, APPLIED ENERGY, 2023

[73] Dong, Hanjiang; Zhu, Jizhong; Li, Shenglin; Wu, Wanli; Zhu, Haohao; Fan, Junwei, Short-term residential household reactive power forecasting considering active power demand via deep Transformer sequence-to-sequence networks, APPLIED ENERGY, 2023

[74] Liu, Jingxuan; Zang, Haixiang; Cheng, Lilin; Ding, Tao; Wei, Zhinong; Sun, Guoqiang, A Transformer-based multimodal-learning framework using sky images for ultra-short-term solar irradiance forecasting, APPLIED ENERGY, 2023

[75] Faiz, Muhammad Faizan; Sajid, Muhammad; Ali, Sara; Javed, Kashif; Ayaz, Yasar, Energy modeling and predictive control of environmental quality for building energy management using machine learning, ENERGY FOR SUSTAINABLE DEVELOPMENT, 2023

[76] Gao, Shuhua; Liu, Yuanbin; Jing, Wang; Wang, Zhengfang; Xu, Wenjun; Yue, Renfeng; Cui, Ruipeng; Yong, Liu; Fan, Xuezhong, Short-term residential load forecasting via transfer learning and multi-attention fusion for EVs' coordinated charging, INTERNATIONAL JOURNAL OF ELECTRICAL POWER & ENERGY SYSTEMS, 2025

[77] Tiwari, Shamik; Jain, Anurag; Ahmed, Nada Mohamed Osman Sid; Charu; Alkwai, Lulwah M.; Dafhalla, Alaa Kamal Yousif; Hamad, Sawsan Ali Saad Machine learning-based model for prediction of power consumption in smart grid- smart way towards smart city
[78] Abdel-Basset, Mohamed; Hawash, Hossam; Chakrabortty, Ripon K.; Ryan, Michael Energy-Net: A Deep Learning Approach for Smart Energy Management in IoT-Based Smart Cities
[79] Nazir, Amril; Shaikh, Abdul Khalique; Shah, Abdul Salam; Khalil, Ashraf, Forecasting energy consumption demand of customers in smart grid using Temporal Fusion Transformer (TFT)
[80] Lee, Jooseung; Kang, Jimyung; Lee, Soonwoo; Oh, Hui-Myoung, Ultra-Short Term Photovoltaic generation Forecasting Based on Data Decomposition and Customized Hybrid Model Architecture
[81] Ma, Qing; Zhang, Deyuan; Wu, Hao; Zhang, Yusen; Yang, Qingrong; Yang, Ning, Shape-Aware Nonintrusive Load Disaggregation via Prior-Guided State-Reused Multibranch Cross-Transformer
[82] Bao, Gengyi; Liu, Xinhua; Zou, Bosong; Yang, Kaiyi; Zhao, Junwei; Zhang, Lisheng; Chen, Muyang; Qiao, Yuanting; Wang, Wentao; Tan, Rui; Wang, Xiangwen, Collaborative framework of Transformer and LSTM for enhanced state-of-charge estimation in lithium-ion batteries
[83] Zhang, Zongbin; Huang, Xiaoqiao; Li, Chengli; Cheng, Feiyan; Tai, Yonghang, CRAformer: A cross-residual attention transformer for solar irradiation multistep forecasting
[84] Yin, Linfei; Zheng, Da, Hybrid modeling with data enhanced driven learning algorithm for smart generation control in multi-area integrated energy systems with high proportion renewable energy
[85] Shan, Zihan; Si, Gangquan; Qu, Kai; Wang, Qianyue; Kong, Xiangguang; Tang, Yu; Yang, ChenMultiscale Self-Attention Architecture in Temporal Neural Network for Nonintrusive Load Monitoring
[86] Bashir, Tasarruf; Wang, Huifang; Tahir, Mustafa; Zhang, Yixiang, Wind and solar power forecasting based on hybrid CNN-ABiLSTM, CNN-transformer-MLP models
[87] Song, Dongran; Rehman, Muhammad Shams Ur; Deng, Xiaofei; Xiao, Zhao; Noor, Javeria; Yang, Jian; Dong, Mi, Accurate solar power prediction with advanced hybrid deep learning approach
[88] Liang, Xuefeng; Hu, Zetian; Zhang, Jun; Chen, Han; Gu, Qingshui; You, Xiaochuan, Developing a robust wind power forecasting method: Integrating data repair, feature screening, and economic impact analysis for practical applications
[89] Lara-Benitez, Pedro; Carranza-Garcia, Manuel; Luna-Romera, Jose Maria; Riquelme, Jose C., Short-term solar irradiance forecasting in streaming with deep learning
[90] Wang, Sicheng; Huang, Yan, Spatio-temporal photovoltaic prediction via a convolutional based hybrid network
[91] Cao, Yizhi; Liao, Yilin; Liu, Zhaoran; Ma, Xiang; Liu, Xinggao, SWAformer: A novel shifted window attention Transformer model for accurate power distribution prediction
[92] Thiyagarajan, Anushalini; Revathi, B. Sri; Suresh, Vishnu, A Deep Learning Model Using Transformer Network and Expert Optimizer for an Hour Ahead Wind Power Forecasting
[93] Mirza, Adeel Feroz; Shu, Zhaokun; Usman, Muhammad; Mansoor, Majad; Ling, Qiang, Quantile-transformed multi-attention residual framework (QT-MARF) for medium-term PV and wind power prediction
[94] Zhu, Leyang; Huang, Xiaoqiao; Zhang, Zongbin; Li, Chengli; Tai, Yonghang, A novel U-LSTM-AFT model for hourly solar irradiance forecasting
[95] Mo, Fan; Jiao, Xuan; Li, Xingshuo; Du, Yang; Yao, Yunting; Meng, Yuxiang; Ding, Shuye, A novel multi-step ahead solar power prediction scheme by deep learning on transformer structure
[96] Jiang, Canchen; Liebman, Ariel; Jie, Bo; Wang, Hao, Dynamic rolling horizon optimization for network-constrained V2X value stacking of electric vehicles under uncertainties
[97] Liao, Xuan; Wong, Man Sing; Zhu, Rui, Dual-gate Temporal Fusion Transformer for estimating large-scale land surface solar irradiation
[98] Wang, Boyu; Dabbaghjamanesh, Morteza; Kavousi-Fard, Abdollah; Yue, Yuntao, AI-enhanced multi-stage learning-to-learning approach for secure smart cities load management in IoT networks
[99] Kim, Jimin; Obregon, Josue; Park, Hoonseok; Jung, Jae-Yoon, Multi-step photovoltaic power forecasting using transformer and recurrent neural networks
[100] Feng, Zhengyuan; Sun, Yuheng; Ning, Jun; Tang, Shoujuan; Liu, Guangxin; Liu, Fangtao; Li, Yang; Shi, Lei, Implementing a provincial-level universal daily industrial carbon emissions prediction by fine-tuning the large language model, APPLIED ENERGY, 2025

[101]     Wang, Ru-Guan; Chuang, Mei-Ling; Ke, Chi-Yun; Chien, Yi-Fan; Ho, Wen-Jen; Chiang, Kuei-Chun; Hung, Yung-Chieh; Tsai, Chu Hsien; Chou, Chien-Cheng, Predicting Imminent Electrical Safety Incidents Using Smart Meter Big Data With Large Language Models, IEEE ACCESS, 2024
[102]     Choi, Sebin; Yoon, Sungmin, GPT-based data-driven urban building energy modeling (GPT-UBEM): Concept, methodology, and case studies, ENERGY AND BUILDINGS, 2024
[103]     Parizad, A.; Hatziadoniu, C. J., Using Prophet Algorithm for Pattern Recognition and Short Term Forecasting of Load Demand Based on Seasonality and Exogenous Features, 2020 52ND NORTH AMERICAN POWER SYMPOSIUM (NAPS), 2021
[104]     Shahin, Matin; Simjoo, Mohammad, Potential applications of innovative AI-based tools in hydrogen energy development: Leveraging large language model technologies, INTERNATIONAL JOURNAL OF HYDROGEN ENERGY, 2025
[105]     Liao, Wenlong; Wang, Shouxiang; Yang, Dechang; Yang, Zhe; Fang, Jiannong; Rehtanz, Christian; Porte-Agel, Fernando, TimeGPT in load forecasting: A large time series model perspective, APPLIED ENERGY, 2025
[106]     Zhang, Liang; Ford, Vitaly; Chen, Zhelun; Chen, Jianli, Automatic building energy model development and debugging using large language models agentic workflow, ENERGY AND BUILDINGS, 2025
[107]     Qaisar, Irfan; Liang, Wei; Sun, Kailai; Xing, Tian; Zhao, Qianchuan, An experimental comparative study of energy saving based on occupancy-centric control in smart buildings, BUILDING AND ENVIRONMENT, 2025
[108]     Zhang, Liang; Chen, Zhelun; Ford, Vitaly, Advancing building energy modeling with large language models: Exploration and case studies, ENERGY AND BUILDINGS, 2024
[109]     Liu, Changchun; Nie, Qingwei, A blockchain-based LLM-driven energy-efficient scheduling system towards distributed multi-agent manufacturing scenario of new energy vehicles within the circular economy, COMPUTERS & INDUSTRIAL ENGINEERING, 2025
[110]     Zhao, Tianqiao; Yogarathnam, Amirthagunaraj; Yue, Meng, A Large Language Model for Determining Partial Tripping of Distributed Energy Resources, IEEE TRANSACTIONS ON SMART GRID, 2025
[111]     Buster, Grant; Pinchuk, Pavlo; Barrons, Jacob; McKeever, Ryan; Levine, Aaron; Lopez, Anthony, Supporting energy policy research with large language models: A case study in wind energy siting ordinances, ENERGY AND AI, 2024
[112]     Liu, Qiang; Mu, Junsheng; Chen, Da; Zhang, Ronghui; Liu, Yijian; Hong, Tao, LLM Enhanced Reconfigurable Intelligent Surface for Energy-Efficient and Reliable 6G IoV, IEEE TRANSACTIONS ON VEHICULAR TECHNOLOGY, 2025
[113]     Xu, Yifang; Zhu, Siyao; Cai, Jiannan; Chen, Jianli; Li, Shuai, A large language model-based platform for real-time building monitoring and occupant interaction, JOURNAL OF BUILDING ENGINEERING, 2025
[114]     Ma, Yichuan X.; Yeung, Lawrence K., BEForeGAN: An image-based deep generative approach for day-ahead forecasting of building HVAC energy consumption, APPLIED ENERGY, 2024
[115]     Asimopoulos, Dimitrios Christos; Radoglou-Grammatikis, Panagiotis; Makris, Ioannis; Mladenov, Valeri; Psannis, Konstantinos E.; Goudos, Sotirios; Sarigiannidis, Panagiotis, Breaching the Defense: Investigating FGSM and CTGAN Adversarial Attacks on IEC 60870-5-104 AI-enabled Intrusion Detection Systems, 18TH INTERNATIONAL CONFERENCE ON AVAILABILITY, RELIABILITY & SECURITY, ARES 2023, 2023
[116]     Nie, Yuhao; Zelikman, Eric; Scott, Andea; Paletta, Quentin; Brandt, Adam, SkyGPT: Probabilistic ultra-short-term solar forecasting using synthetic sky images from physics-constrained VideoGPT, ADVANCES IN APPLIED ENERGY, 2024
[117]     Heo, SungKu; Ko, Jaerak; Kim, SangYoun; Jeong, Chanhyeok; Hwangbo, Soonho; Yoo, ChangKyoo, Explainable AI-driven net-zero carbon roadmap for petrochemical industry considering stochastic scenarios of remotely sensed offshore wind energy, JOURNAL OF CLEANER PRODUCTION, 2022
[118]     Zhang, L., & Chen, Z. (2025). Opportunities of applying Large Language Models in building energy sector. Renewable and Sustainable Energy Reviews, 214, 115558.
[119]     Ullah, A., Qi, G., Hussain, S., Ullah, I., & Ali, Z. (2024). The role of llms in sustainable smart cities: Applications, challenges, and future directions. arXiv preprint arXiv:2402.14596.
[120]     Raffel, C., Shazeer, N., Roberts, A., Lee, K., Narang, S., Matena, M., ... & Liu, P. J. (2020). Exploring the limits of transfer learning with a unified text-to-text transformer. Journal of machine learning research, 21(140), 1-67.
[121]     Ruan, J., Liang, G., Zhao, H., Liu, G., Sun, X., Qiu, J., ... & Dong, Z. Y. (2024). Applying large language models to power systems: potential security threats. IEEE transactions on smart grid.

[122] Devlin, J., Chang, M. W., Lee, K., & Toutanova, K. (2019, June). Bert: Pre-training of deep bidirectional transformers for language understanding. In Proceedings of the 2019 conference of the North American chapter of the association for computational linguistics: human language technologies, volume 1 (long and short papers) (pp. 4171-4186).

[123] Jiang, G., Ma, Z., Zhang, L., & Chen, J. (2024). EPlus-LLM: A large language model-based computing platform for automated building energy modeling. Applied Energy, 367, 123431.

[124] Acharya, D. B., Kuppan, K., & Divya, B. (2025). Agentic AI: Autonomous Intelligence for Complex Goals–A Comprehensive Survey. IEEE Access

[125] Zihang Qiu, Chaojie Li, Zhongyang Wang, Huadong Mo, Renyou Xie, Guo Chen, Zhaoyang Dong, FPE-LLM: Highly Intelligent Time-Series Forecasting and Language Interaction LLM in Energy Systems, arXiv:2411.00852v1 [cs.LG] 30 Oct 2024

[126] Arcas, G.I.; Cioara, T.; Anghel, I.; Lazea, D.; Hangan, A. Edge Offloading in Smart Grid. Smart Cities 2024, 7, 680-711.

[127] Seyyedreza Madani, Ahmadreza Tavasoli, Zahra Khoshtarash Astaneh, Pierre-Olivier Pineau, Large Language Models integration in Smart Grids, arXiv:2504.09059v1 [cs.CY] 12 Apr 2025

[128] Gao, C., Lan, X., Li, N. et al. Large language models empowered agent-based modeling and simulation: a survey and perspectives. Humanit Soc Sci Commun 11, 1259 (2024). https://doi.org/10.1057/s41599-024-03611-3

[129] Opy Das, Muhammad Hamza Zafar, Filippo Sanfilippo, Souman Rudra, Mohan Lal Kolhe, Advancements in digital twin technology and machine learning for energy systems: A comprehensive review of applications in smart grids, renewable energy, and electric vehicle optimisation, Energy Conversion and Management: X, Volume 24, 2024, 100715, ISSN 2590-1745, https://doi.org/10.1016/j.ecmx.2024.100715.

[130] Muhammad Arslan, Lamine Mahdjoubi, Saba Munawar, Driving sustainable energy transitions with a multi-source RAG-LLM system, Energy and Buildings, Volume 324, 2024, 114827, ISSN 0378-7788, https://doi.org/10.1016/j.enbuild.2024.114827.

[131] Aske Plaat, Max van Duijn, Niki van Stein, Mike Preuss, Peter van der Putten, Kees Joost Batenburg, Agentic Large Language Models, a survey, arXiv:2503.23037v2 [cs.AI] 03 Apr 2025

[132] Andrea Matarazzo, Riccardo Torlone, A Survey on Large Language Models with some Insights on their Capabilities and Limitations, arXiv:2501.04040v1 [cs.CL] 03 Jan 2025

[133] Thapa, S., Shiwakoti, S., Shah, S.B. et al. Large language models (LLM) in computational social science: prospects, current state, and challenges. Soc. Netw. Anal. Min. 15, 4 (2025). https://doi.org/10.1007/s13278-025-01428-9

[134] Liu, M., Zhang, L., Chen, J. et al. Large language models for building energy applications: Opportunities and challenges. Build. Simul. 18, 225–234 (2025).

[135] Linyao Yang, Shi Luo, Xi Cheng, Lei Yu, Leveraging Large Language Models for Enhanced Digital Twin Modeling: Trends, Methods, and Challenges, arXiv:2503.02167v1 [cs.ET] 04 Mar 2025